\RequirePackage{fix-cm}
\documentclass{svjour3}
\smartqed
\usepackage{graphicx}
\usepackage{multirow}
\usepackage{multicol}
\usepackage{array}
\usepackage{enumitem}
\usepackage{cite}
\usepackage{booktabs}
\usepackage{bm}
\usepackage{comment}
\usepackage{hyperref}
\hypersetup{colorlinks,
    linkcolor=blue,
    citecolor=blue,
    urlcolor=blue}
\usepackage{amsmath,amssymb,amsfonts}
\usepackage{mathtools}
\usepackage{bbding}
\usepackage{subcaption}
\usepackage{algorithm}
\usepackage{algpseudocode}
\usepackage{float}
\usepackage{orcidlink}
\usepackage{url}

\usepackage{pifont}
\usepackage[outline]{contour}
\newcommand{\cmark}{\contour{black}{\textcolor{green}{\ding{51}}}}
\newcommand{\xmark}{\contour{black}{\textcolor{red}{\ding{55}}}}

\newcommand{\R}{\mathbb{R}}

\newcommand{\Ind}{\mathbf{1}}
\newcommand{\Conv}{\mathrm{Conv}}
\newcommand{\GConv}{\mathrm{Conv}_G}
\newcommand{\GCLIF}{\textsc{GC-LIF}}
\newcommand{\SGE}{\textsc{SGE-Net}}
\newcommand{\rot}{\rho}

\makeatletter
\@fleqnfalse
\makeatother

\begin{document}
\journalname{Preprint}

\title{GC-LIF: Spiking Group-Equivariant Neural Network With Group-Coupled Membrane Dynamics}
%\begin{comment}
\author{Nhan Trong Luu$*$ \orcidlink{0009-0000-6783-6257} \and Duong Trung Luu \orcidlink{0009-0004-6714-2270} \and  Truong Cong Thang \orcidlink{0000-0002-0923-1927}}
\institute{ 
        Nhan Trong Luu \at College of ICT, Can Tho University, Can Tho, Vietnam. \email{luutn@ctu.edu.vn}
        \and
        Duong Trung Luu \at Center for Digital Transformation and Communication, Can Tho University, Can Tho, Vietnam. \email{luutd@ctu.edu.vn}
        \and
        Truong Cong Thang \at Department of Computer Science and Engineering, The University of Aizu, Japan. \email{thang@u-aizu.ac.jp}
        \and
        $*$Corresponding author.
}
%\end{comment}
%\author{Anonymous authors}

\date{Revised \today}
\maketitle
\begin{abstract}
Spiking neural networks (SNNs) promise event-driven, low-power inference, but on
vision tasks they still acquire rotation invariance only through data
augmentation. Group-equivariant CNNs build such symmetries into the
architecture, yet a faithful spiking analogue is missing: a leaky
integrate-and-fire (LIF) neuron placed after a $G$-equivariant convolution can
break equivariance through its membrane recurrence. We give a necessary and
sufficient condition for this not to happen. The layer is equivariant at every
time step if and only if the neuron parameters $(\beta, v_{\mathrm{th}})$ are
constant along the joint orbits of the group action on the (group, space)
index. We solve it in closed form, and it covers any elementwise stateful
neuron. Its sufficiency half specialises a
known result for equivariant recurrent networks; its necessity half rules out
the heterogeneous per-neuron parameters common in SNNs. The per-channel
parameterisation that satisfies it, Group-Coupled LIF (GC-LIF), is combined with
$p4$ and $p4m$-equivariant convolutions and a spike-rate group max-pool into
SGE-Net. On MNIST, KMNIST, Fashion-MNIST, CIFAR-10 and CIFAR-100, evaluating
each checkpoint on the clean test set and all its group-transformed copies,
SGE-Net trained without rotations is invariant under
$C_4$ and $D_4$ and outperforms a plain SNN trained with rotations in nine of ten
dataset-group pairs, whereas non-equivariant baselines lose up to $72$
percentage points in the worst case. Operator-level tests confirm the theory
before and after training, and an analytical estimate places SGE-Net at
$1.7\times$ the arithmetic energy of a standard CNN and $2.3\times$ below its
ANN counterpart at eight time steps.
\end{abstract}

Convolutional neural networks (CNNs)~\cite{lecun1998gradient} are
translation-equivariant by construction, which is a major reason for their
sample efficiency. Group-equivariant CNNs (G-CNNs)~\cite{cohen2016group} and
steerable CNNs~\cite{cohen2016steerable, weiler2019general} extend this to
rotations and reflections, and improve parameter efficiency and generalisation
when data arrive in arbitrary poses, as in medical imaging, microscopy and
satellite imagery.

Spiking neural networks (SNNs) have meanwhile become a practical substrate for
low-power, event-driven inference, and surrogate-gradient
training~\cite{neftci2019surrogate, luu2026survey} brings them close to ANN
accuracy on vision benchmarks. On rotation-sensitive tasks, however, SNNs still
rely on data augmentation~\cite{shen2023training} or fixed transformation
priors~\cite{pedersen2025covariant} rather than on the exact discrete-group
equivariance of G-CNNs.

Simply inserting a spiking neuron into a G-CNN does not solve this. Equivariance
requires every internal feature to transform along with the input, and a spiking
neuron carries a membrane state across time steps. If its parameters differ
between the positions that the group action exchanges, the rotated and
unrotated networks drift apart step by step, even though every convolution in
the stack is equivariant. This paper works out exactly which parameterisations
avoid this and what the resulting network buys and costs. We make four
contributions:
\begin{itemize}
    \item \textbf{A necessary and sufficient condition.} A $G$-equivariant
    convolution followed by per-element LIF neurons is equivariant at every time
    step if and only if $(\beta, v_{\mathrm{th}})$ are constant along the joint
    orbits of the group action on the (group, space) index. We solve this
    constraint in closed form and show that it covers any elementwise stateful
    neuron. The sufficiency half is the spiking instance of a known induction
    for equivariant recurrent networks~\cite{keller2025flow}; the necessity
    half, and the list of common SNN parameterisations it excludes, are
    specific to this work.
    \item \textbf{GC-LIF and SGE-Net.} Group-Coupled LIF (\GCLIF{}) keeps one
    membrane state per (group, spatial) location but shares $(\beta,
    v_{\mathrm{th}})$ per channel, the only choice that is both orbit invariant
    and translation equivariant. With lifting and group convolutions, group
    BatchNorm and a spike-rate group max-pool it forms the Spiking Group
    Equivariant Network (\SGE{}).
    \item \textbf{A controlled evaluation.} We compare \SGE{} with a plain CNN, a
    plain SNN and an ANN G-CNN on five datasets, four groups ($C_4$, $C_8$,
    $D_4$, $D_8$) and three training and test regimes. Every number comes from
    one checkpoint, selected on a held-out validation split and evaluated on the
    clean test set and on every group-transformed copy of it, so that exact
    invariance can be read off the tables directly. We report $95\%$ confidence
    intervals over four seeds, a sensitivity study over the number of time
    steps $T$, and an analytical energy estimate.
    \item \textbf{Operator-level validation.} We test the theory on the
    operators themselves, at initialisation and after training: tied parameters
    give identical spike trains under $C_4$ and $D_4$ up to float32 rounding,
    untying them breaks this immediately, and the threshold and the leak break
    it through different mechanisms, exactly as the proof predicts.
\end{itemize}

To our knowledge, \SGE{} is the first spiking network with exact discrete-group
equivariance through time, exact for $n \in \{1,2,4\}$, where the group acts by
pixel permutations, and approximate for $n = 8$. Equivariance of recurrent
networks under static group actions is itself established
theory~\cite{keller2025flow, azari2022equivariant, nguyen2023equivariant}, and
we do not claim it. Our contribution is the characterisation of which spiking
parameterisations satisfy it and a controlled measurement of the
resulting accuracy and cost.

\section{Related Work}

\subsection{Group-equivariant CNNs}
Cohen and Welling~\cite{cohen2016group} introduced G-CNNs for discrete cyclic and dihedral groups.
Steerable CNNs~\cite{cohen2016steerable, weiler2019general} generalise them to
continuous groups through irreducible
representations~\cite{bekkers2019b, finzi2020generalizing}. Later work covers
separable G-CNNs~\cite{knigge2022exploiting}, partial
equivariance~\cite{romero2022learning, kim2024variational} and steerable
transformers~\cite{kundu2025steerable}.

\subsection{Spiking neural networks for vision}
Surrogate-gradient training~\cite{neftci2019surrogate,
zenke2018superspike, fang2021deep} gives strong SNN results on MNIST, CIFAR-10
and event-based data such as DVS-CIFAR10~\cite{luu2026survey, luu2024,
luu2026parameter, luu2025hybrid}. Recent work addresses long-range
dependencies~\cite{du2024spiking} and convolutional
backbones~\cite{liao2025spikeatconv}. All of these treat geometric invariance
as an augmentation problem.

\subsection{Equivariance in recurrent and sequence models}
\label{sec:rw_rnn}
Equivariance of stateful models under static group actions is well understood.
Keller~\cite{keller2025flow} studies equivariance of sequence models with respect to
time-parameterised flows and, on the way, proves by induction that a recurrent
network built from a group-equivariant input map, a group-equivariant recurrent
map, a pointwise equivariant nonlinearity and a group-invariant initial state is
equivariant at every time step. Static equivariance has also been built into
recurrent dynamical models for motion prediction~\cite{azari2022equivariant} and
recurrent policies for partially observed control~\cite{nguyen2023equivariant},
and the principle that compositions of equivariant maps are equivariant goes
back to~\cite{kondor2018generalization}. The sufficiency half of
Theorem~\ref{thm:main} is the spiking instance of this line of work
(Section~\ref{sec:relation_rnn}). The architectures in that literature carry no
per-element parameters, so the question of which parameterisations of a
stateful unit preserve equivariance does not arise there. It does arise for
SNNs, where per-neuron time constants and thresholds are
routine~\cite{fang2021incorporating}, and Proposition~\ref{prop:naive_fails}
answers it.

\subsection{Symmetries in SNNs}
Pedersen et al.~\cite{pedersen2025covariant} build covariant spatio-temporal receptive fields
for event-based vision from fixed Gaussian kernels, covering spatial affine and
temporal scaling transformations but not learnable discrete rotation and
reflection groups. Pedersen et al.~\cite{pedersen2026scale} use the scale covariance of the LIF
membrane to build spiking wavelets; the symmetry there is temporal scale, and
the kernels are fixed. Augmentation-based methods~\cite{shen2023training,
shen2023robust} obtain approximate rotation robustness through preprocessing. We
are not aware of prior work stating when the membrane recurrence itself commutes
with a discrete spatial group action.

\section{Theoretical basis}

\subsection{Background}

\subsubsection{LIF neurons}
A discrete-time leaky integrate-and-fire neuron has membrane potential $v$, leak
factor $\beta \in (0,1)$ and threshold $v_{\mathrm{th}}$. With reset to zero
after a spike,
\begin{equation}
\label{eq:lif}
v_t = \beta\, v_{t-1}\,(1 - s_{t-1}) + i_t, \quad
s_t = \Ind\left[v_t \geq v_{\mathrm{th}}\right].
\end{equation}
Training uses an arctangent surrogate gradient for the indicator
(Section~\ref{sec:arch_gclif}).

\subsubsection{Feature maps and the group action}
We work with the cyclic group $G = C_n = \{g_0, \dots, g_{n-1}\}$, where $g_a$
is a planar rotation by $\theta_a := 2\pi a/n$ and $g_a g_b = g_{a+b \bmod n}$.
All arithmetic on group indices is modulo $n$. The dihedral case is analogous
and is detailed in Appendix~\ref{apd:dihedral_conv}.

The following definition fixes the index notation used in every statement
below.

\begin{definition}[Feature maps]
A feature map at a hidden layer is a function
$F : \{0, \dots, n-1\} \times \mathcal{P} \to \R^C$ indexed by a group
coordinate $h$, a spatial coordinate $p \in \mathcal{P} \subseteq \R^2$ and a
channel $c$. We write $F[c, h, p]$, so one time step of a layer is a tensor in
$\R^{C \times n \times |\mathcal{P}|}$.
\end{definition}

Every equivariance statement in this paper is made with respect to one operator
on such feature maps, which we now define.

\begin{definition}[Group action]
For $g_a \in G$ the operator $\rot(g_a)$ acts on a feature map by
\begin{equation}
\label{eq:groupaction}
\left(\rot(g_a) F\right)[c, h, p]
=
F\left[\,c, h - a, R_{-\theta_a}\, p\,\right],
\end{equation}
where $R_\theta$ is the $2\times2$ rotation matrix. It cyclically shifts the
group axis by $a$ and rotates the spatial axes by $\theta_a$. For
$n \in \{1,2,4\}$ it is a permutation of the pixel grid; it only relabels each
position $(h,p)$ as $(h-a, R_{-\theta_a}p)$.
\end{definition}

A map $L$ is \emph{$G$-equivariant} if $L(\rot(g)\,x) = \rot(g)\,L(x)$ for all
$g$ and $x$. A \emph{lifting convolution} maps $\R^2 \to G \times \R^2$ and a
\emph{group convolution} maps $G \times \R^2 \to G \times \R^2$; both tie
weights through rotated copies of a base kernel~\cite{cohen2016group}. The
theorem below treats the convolution as a black box that is equivariant, so we
first state that property together with the sign convention it depends on,
which is easy to get wrong.

\begin{lemma}[Equivariance of group convolution]
\label{lem:conv}
Define the group convolution $\GConv : F \mapsto F \star w$ with kernel
$w[c', c, h', q]$ by
\begin{equation}
\label{eq:gconv}
(F \star w)[c', h, p] = \sum_{c=1}^{C}\sum_{h'=0}^{n-1}\sum_{q \in \mathcal{Q}}
F\left[c, h-h', p - R_{\theta_h} q\right] \cdot w[c', c, h', q],
\end{equation}
where $\mathcal{Q}$ is the kernel footprint. Then
$\GConv\left(\rot(g_a) F\right) = \rot(g_a)\,\GConv(F)$ for every $g_a \in G$
and every $F$.
\end{lemma}
\begin{proof}
See Appendix~\ref{proof:lem:conv}.
\qed\end{proof}

The offset rotation $R_{+\theta_h}$ in Equation~\ref{eq:gconv} must match the
convention of Equation~\ref{eq:groupaction}: since
$R_{-\theta_a}R_{\theta_h} = R_{\theta_{h-a}}$, the spatial rotation then follows
the shifted group index. With $R_{-\theta_h}$ instead, the composition gives
$R_{-\theta_{h+a}}$ and equivariance fails for every $a$ with $2a \not\equiv 0
\bmod n$; Section~\ref{sec:numval} confirms this numerically.

\subsection{Why a naive spiking G-CNN fails through time}

Suppose the LIF parameters $\beta_{c,h,p}$ and $v_{\mathrm{th},c,h,p}$ may vary
over the whole feature index. The group action moves the current that the
unrotated network had at $(h-a, R_{-\theta_a}p)$ to position $(h,p)$. If the
parameters at these two positions differ, the neuron at $(h,p)$ integrates the
right current with the wrong leak or compares it with the wrong threshold, and
its spike train no longer matches the relabelled spike train of the unrotated
network. Because the membrane carries this mismatch to the next step, the error
also accumulates in time. The natural fix, making the parameters independent of
the group index $h$, is neither sufficient nor necessary, because the action
moves $p$ as well as $h$. The correct condition, derived next, is invariance
along the joint orbits of $G$ on $(h,p)$.

\subsection{Condition for space-time equivariance}

Each step of the induction behind Theorem~\ref{thm:main} moves $\rot(g_a)$
past one elementary operation of the LIF update. The next lemma lists exactly
those operations; property (R5) uses Definition~\ref{def:orbit} below.

\begin{lemma}[Properties of $\rot(g_a)$]
\label{lem:rho}
For any $g_a \in G$, feature maps $F, F'$ and scalar $\alpha \in \R$:
\begin{enumerate}[leftmargin=1cm]
    \item[(R1)] \textbf{Scalar multiplication:}
                $\rot(g_a)(\alpha F) = \alpha \left(\rot(g_a) F\right)$.
    \item[(R2)] \textbf{Addition:}
                $\rot(g_a)(F + F') = \rot(g_a) F + \rot(g_a) F'$.
    \item[(R3)] \textbf{Hadamard product:}
                $\rot(g_a)(F \odot F') = \rot(g_a) F \odot \rot(g_a) F'$.
    \item[(R4)] \textbf{Pointwise function} $\phi : \R \to \R$:
                $\rot(g_a)(\phi(F)) = \phi(\rot(g_a) F)$.
    \item[(R5)] \textbf{Orbit-invariant scaling:} if a parameter field
                $\lambda$ satisfies Equation~\ref{eq:orbit}, then
                $\rot(g_a)(\lambda \odot F) = \lambda \odot \rot(g_a) F$.
                (R1) is the special case of a constant field.
\end{enumerate}
\end{lemma}
\begin{proof}
See Appendix~\ref{proof:lem:rho}.
\qed\end{proof}

\subsubsection{Orbit-invariant parameter fields}
\label{sec:orbit}

The theorem's hypothesis is membership of the following class of parameter
fields, so we need to name it. It is strictly larger than, and differently
shaped from, the class of fields that do not depend on $h$.

\begin{definition}[Orbit-invariant parameter field]
\label{def:orbit}
A parameter field $\lambda = (\lambda_{c,h,p})$ is \emph{orbit invariant} if
\begin{equation}
\label{eq:orbit}
\lambda_{c,h,p} = \lambda_{c,\,h-a,\,R_{-\theta_a} p}
\qquad \text{for every } c,\, h,\, p,\, a,
\end{equation}
that is, if $\lambda$ is constant on each orbit of the joint action
$(h,p) \mapsto (h-a, R_{-\theta_a} p)$.
\end{definition}

Equation~\ref{eq:orbit} quantifies over all $a$, which makes it awkward to
check. The next lemma replaces it with an explicit parameterisation, so that
one can tell at a glance whether a given design is admissible.

\begin{lemma}[Characterisation of orbit-invariant fields]
\label{lem:orbit}
A field $\lambda$ is orbit invariant if and only if there are functions
$f_c : \mathcal{P} \to \R$, one per channel, such that
\begin{equation}
\label{eq:orbit_solution}
\lambda_{c,h,p} = f_c\!\left(R_{-\theta_h} p\right)
\qquad \text{for every } c,\,h,\,p.
\end{equation}
\end{lemma}
\begin{proof}
See Appendix~\ref{proof:lem:orbit}.
\qed\end{proof}

Four consequences follow and delimit what a designer may do:
\begin{enumerate}[leftmargin=1cm]
    \item[(C1)] \textbf{Spatially homogeneous fields.} If $\lambda_{c,h,p}
    \equiv \lambda_{c,h}$, then $f_c$ must be constant, so the condition reduces
    to ``$\lambda_{c,h}$ does not depend on $h$''. Only in this case is the
    familiar statement correct.
    \item[(C2)] \textbf{Independence of $h$ is not sufficient.} If
    $\lambda_{c,h,p} \equiv \lambda_{c,p}$, the condition becomes
    $\lambda_{c,p} = \lambda_{c,R_{-\theta_a}p}$ for every $a$: $\lambda_c$ must
    be constant on each $C_n$-orbit of pixel positions. A per-neuron learnable
    time constant, which varies freely over $p$, violates it.
    \item[(C3)] \textbf{Independence of $h$ is not necessary.} Any non-constant
    $f_c$ in Equation~\ref{eq:orbit_solution} gives an orbit-invariant field
    that depends on $h$: slot $h$ receives the profile of slot $0$ rotated by
    $\theta_h$.
    \item[(C4)] \textbf{\GCLIF{} is the intersection.} A field that is orbit
    invariant and spatially homogeneous is a per-channel constant $\lambda_c$.
    Spatial homogeneity is what translation equivariance requires, so
    per-channel constants are the only parameters that are both
    $C_n$-equivariant and translation-equivariant. \GCLIF{} uses them.
\end{enumerate}

\subsubsection{The space-time equivariance theorem}

The following theorem is what allows a stateful spiking neuron to sit inside an
equivariant stack at all: without it, a network built only from equivariant
convolutions can still lose equivariance once a neuron with memory is inserted,
and a mismatched leak does not even show at the first time step
(Section~\ref{sec:numval}).

\begin{theorem}[Space-time equivariance of LIF over group coupling]
\label{thm:main}
Let $L = \mathrm{LIF} \circ \GConv$ be a $G$-equivariant group convolution
followed by per-element LIF neurons whose parameters $\beta_{c,h,p} \in (0,1)$
and $v_{\mathrm{th},c,h,p} > 0$ are orbit invariant, with dynamics
\begin{equation}
\label{eq:lif_state}
\begin{aligned}
    v_t[c,h,p] &= \beta_{c,h,p}\, v_{t-1}[c,h,p]\left(1 - s_{t-1}[c,h,p]\right) + i_t[c,h,p],\\
    s_t[c,h,p] &= \Ind\left[v_t[c,h,p] \geq v_{\mathrm{th},c,h,p}\right],\qquad v_0 = 0, \quad s_0 = 0,
\end{aligned}
\end{equation}
where $i_t := \GConv(x_t)$. Then for every $g_a \in G$ and every input sequence
$x_{1:T}$,
\begin{equation}
\label{eq:thm_claim}
L\left(\rot(g_a)\, x_{1:T}\right) = \rot(g_a)\, L(x_{1:T}),
\end{equation}
where $\rot(g_a)$ acts on each time step separately:
$(\rot(g_a) x_{1:T})_t := \rot(g_a) x_t$.
\end{theorem}
\begin{proof}
See Appendix~\ref{proof:thm:main}.
\qed\end{proof}

The proof uses orbit invariance at exactly two places: when the leak is moved
through the index relabelling and when the threshold is moved through the
indicator. Everything else is bookkeeping about relabelled coordinates. Two
readings follow. Sharing across channels is never needed, and the algebraic form
of the LIF update plays no role beyond being applied elementwise
(Section~\ref{sec:general_neuron}).

The converse is what constrains design in practice, because it says whether a
heterogeneous parameterisation can be kept or must be abandoned.

\begin{proposition}[Orbit invariance is necessary]
\label{prop:naive_fails}
Let $L^{\mathrm{naive}}$ be as in Theorem~\ref{thm:main} but with
$\beta_{c,h,p}$ and $v_{\mathrm{th},c,h,p}$ unrestricted. Then
$L^{\mathrm{naive}}(\rot(g_a) x_{1:T}) = \rot(g_a) L^{\mathrm{naive}}(x_{1:T})$
for every input and every $g_a \in G$ if and only if both $\beta$ and
$v_{\mathrm{th}}$ are orbit invariant, that is, if and only if both have the
form of Equation~\ref{eq:orbit_solution}. In particular, independence from
$h$ alone is neither sufficient (C2) nor necessary (C3).
\end{proposition}
\begin{proof}
See Appendix~\ref{apd:naive_fails}.
\qed\end{proof}

Together, Theorem~\ref{thm:main} and Proposition~\ref{prop:naive_fails} give a
test for any candidate parameterisation: it is admissible exactly when it
factors through Equation~\ref{eq:orbit_solution}. A per-neuron learnable time
constant, as in Parametric LIF~\cite{fang2021incorporating}, fails the test; a
per-channel one passes; a per-orientation one, which looks natural on a lifted
tensor, fails by (C1).

\subsection{Extension to other spiking neuron models}
\label{sec:general_neuron}

The argument above is easy to mistake for a property of the LIF equation, which
would make it fragile under a change of neuron model. The next proposition
shows it is not: what matters is that the update acts elementwise and that its
parameters are orbit invariant.

\begin{proposition}[Equivariance of elementwise stateful neurons]
\label{prop:general_neuron}
Let $u_t[c,h,p] \in \R^d$ be a neuron state with
\begin{equation}
\label{eq:general_neuron}
u_t[c,h,p] = \Psi\!\left(u_{t-1}[c,h,p],\, i_t[c,h,p];\, \lambda_{c,h,p}\right),
\qquad
s_t[c,h,p] = \Theta\!\left(u_t[c,h,p];\, \lambda_{c,h,p}\right),
\end{equation}
for fixed maps $\Psi : \R^d \times \R \times \R^m \to \R^d$ and
$\Theta : \R^d \times \R^m \to \R$, a parameter field
$\lambda_{c,h,p} \in \R^m$ and an initial state $u_0$ that is constant over
$(h,p)$. Let $L := \mathcal{N} \circ \GConv$, where $\mathcal{N}$ is the
recursion of Equation~\ref{eq:general_neuron}. If $\lambda$ is orbit invariant,
then $L(\rot(g_a) x_{1:T}) = \rot(g_a) L(x_{1:T})$ for every $g_a$ and every
input. If $\Psi$ or $\Theta$ depends non-trivially on $\lambda$, orbit
invariance is also necessary.
\end{proposition}
\begin{proof}
See Appendix~\ref{proof:prop:general}.
\qed\end{proof}

Checking standard neuron models against the two requirements:
\begin{itemize}
    \item \textbf{Covered as is:} LIF with subtractive reset
    ($v_t \mapsto v_t - v_{\mathrm{th}} s_t$); current-based (second-order) LIF
    with state $(v, I)$; adaptive-threshold LIF with state $(v, \vartheta)$
    driven by the neuron's own spikes; the Izhikevich model with state $(v, w)$
    and a quadratic $\Psi$; exponential and quadratic integrate-and-fire; and
    Parametric LIF~\cite{fang2021incorporating} with a per-channel learnable
    time constant.
    \item \textbf{Covered with a constraint:} neurons with lateral coupling, such
    as winner-take-all or inhibition across orientations, are not elementwise.
    They remain admissible if the coupling is itself $G$-equivariant, which for
    inhibition across orientations means a circulant kernel on the group axis.
    \item \textbf{Not covered:} neurons with independent stochastic firing,
    unless the noise is transported by $\rot(g_a)$. For those the guarantee
    weakens to equivariance in distribution.
\end{itemize}

\subsection{Relation to equivariant recurrent network theory}
\label{sec:relation_rnn}

The sufficiency half of Theorem~\ref{thm:main} specialises an established
argument. Keller~\cite{keller2025flow} proves by induction that a recurrent network
assembled from a $G$-equivariant input map, a $G$-equivariant recurrent map, a
pointwise equivariant nonlinearity and a group-invariant initial state is
$G$-equivariant at every step; instantiating it with group convolutions and a
spiking nonlinearity gives our sufficiency direction, as it would for the
recurrent models of~\cite{azari2022equivariant, nguyen2023equivariant}. An
ordinary LIF with a scalar or per-channel leak and threshold already satisfies
the condition, and \GCLIF{} does not make such a neuron more equivariant.

Three points are specific to our setting. First, the LIF update is not of the
form $h_{t+1} = \sigma(\mathcal{W} \star h_t + \mathcal{U} \star f_t)$ assumed
in~\cite{keller2025flow}: the reset multiplies the state by a thresholded
function of itself, so the induction also needs the Hadamard property (R3), and
the nonlinearity is a discontinuous indicator. This is a modest technical
difference. Second, and more substantively, prior results are sufficiency
statements that assume every component is equivariant by construction; they do
not say which parameterisations of a stateful unit are admissible, because the
units they consider carry no per-element parameters.
Proposition~\ref{prop:naive_fails} and Lemma~\ref{lem:orbit} answer that
question constructively, and the answer matters for SNNs, where heterogeneous
parameters are standard. Third, we measure the failure modes
(Section~\ref{sec:numval}) and show that the threshold and the leak break
equivariance through different mechanisms and on different time scales.

\section{SGE-Net: a group-equivariant SNN with \GCLIF{}}
\label{sec:architecture}

\begin{figure*}[t!]
    \centering
    \includegraphics[width=\linewidth]{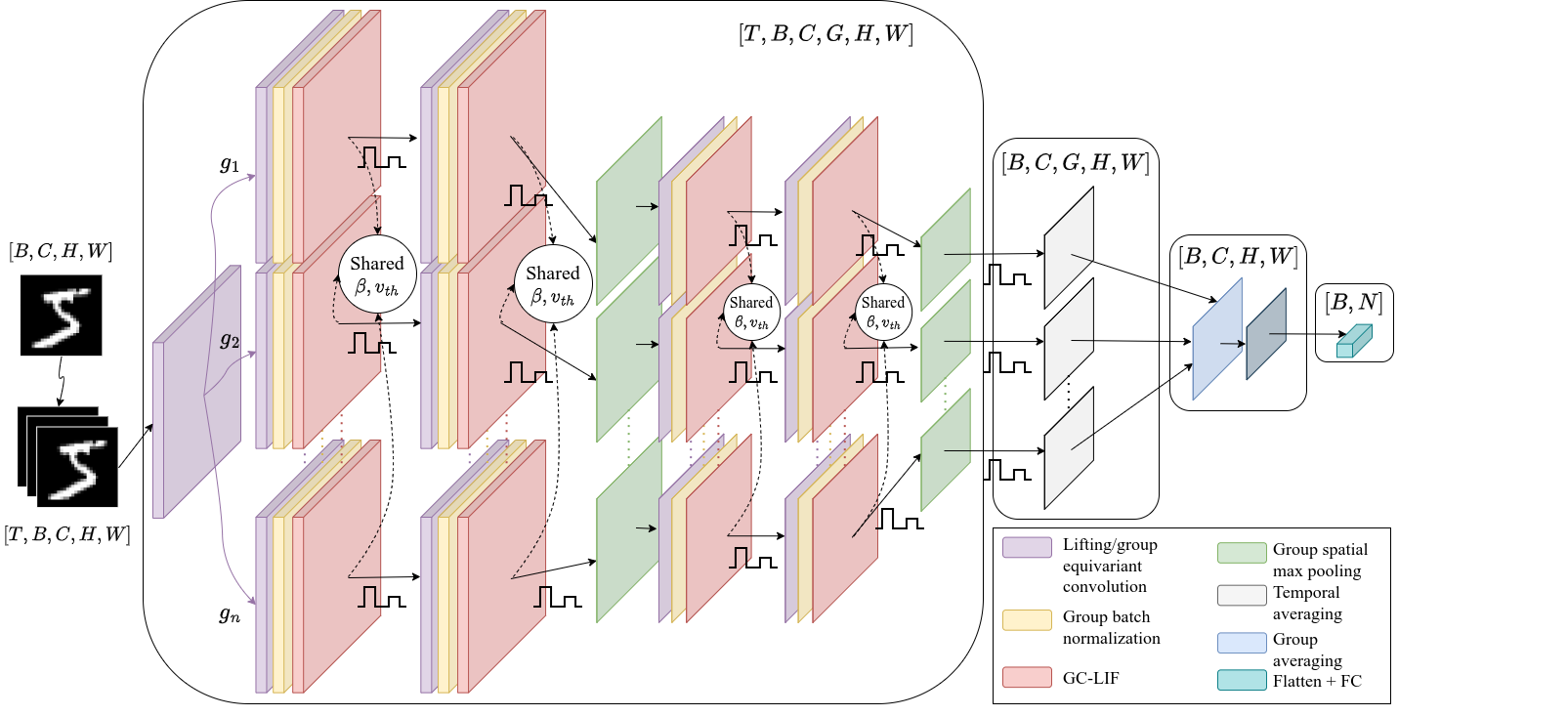}
    \caption{Overview of \SGE{}: lifting and group convolutions, group BatchNorm
    and \GCLIF{} neurons, group spatial average pooling between blocks, and a
    readout that averages spike rates over time and max-pools over the group
    axis.}
    \label{fig:scheme}
\end{figure*}

\SGE{} is a stack of equivariant blocks followed by an invariant readout
(Figure~\ref{fig:scheme}). The input is encoded directly, $x_t = x$ for
$t = 1, \dots, T$, and block $\ell$ applies at every time step
\begin{equation}
\label{eq:block}
i^{(\ell)}_t = \mathrm{BN}^{(\ell)}\left(\Conv^{(\ell)}\, F^{(\ell-1)}_t\right),
\qquad
\left(F^{(\ell)}_t,\, v^{(\ell)}_t\right)=\GCLIF^{(\ell)}\left(i^{(\ell)}_t,\, v^{(\ell)}_{t-1}\right),
\end{equation}
where $\Conv^{(\ell)}$ is the lifting convolution for $\ell = 1$ and a group
convolution for $\ell \geq 2$, $\mathrm{BN}^{(\ell)}$ is group BatchNorm and
$v^{(\ell)}_0 = 0$. Every component commutes with $\rot(g_a)$, as shown below,
so the whole stack does.

\subsection{Lifting convolution}
\label{sec:lift}

The first layer turns an image into a $G$-indexed feature map. For a base kernel
$K \in \R^{C_1 \times C_{\text{in}} \times k \times k}$,
\begin{equation}
\label{eq:lift}
\left(\mathrm{Lift}\, x\right)[c, h, p]=\sum_{q \in \mathcal{N}}\left\langle K_h\left[c, :, q\right], x\left[:,p + q\right]\right\rangle,
\qquad K_h(q) := K\left(R_{-\theta_h}\, q\right),
\end{equation}
so slot $h$ holds the base kernel rotated by $\theta_h$ and $K$ is the only
parameter. The standard property~\cite{cohen2016group, weiler2019general}
\begin{equation}
\label{eq:lift_equiv}
\mathrm{Lift}\left(R_{\theta_a} x\right) = \rot(g_a)\,\mathrm{Lift}(x)
\end{equation}
says that rotating the image shifts which slots respond rather than changing the
response.

For $n \in \{1,2,4\}$ the rotated kernel is a permutation of kernel entries and
is exact. For $n > 4$, bilinear resampling of a non-smooth kernel is inaccurate,
so we parameterise $K$ in a Gaussian-radial, Fourier-angular steerable
basis~\cite{worrall2017harmonic, cohen2016steerable, andrearczyk2019exploring}:
\begin{equation}
\label{eq:steerable}
K(r,\varphi)= \sum_{m,\ell} \psi_m(r)\left(\alpha_{m,\ell}\cos(\ell\varphi)+ \beta_{m,\ell}\sin(\ell\varphi)\right),
\qquad \psi_m(r) := e^{-(r-r_m)^2 / 2\sigma^2},
\end{equation}
with trainable coefficients $(\alpha_{m,\ell}, \beta_{m,\ell})$. Rotation is
then the exact relabelling $K_h(r,\varphi) = K(r,\varphi - \theta_h)$ of the
continuous kernel. The layer, however, uses $K_h$ sampled on the $k \times k$
grid, and sampling does not commute with rotation for angular frequencies the
grid cannot resolve. The basis therefore removes the resampling error but not
the sampling error, and equivariance at $n = 8$ remains approximate
(Section~\ref{sec:numval}).

\subsection{Group BatchNorm}
\label{sec:gbn}

Group BatchNorm pools statistics over the group axis as well as over batch and
space:
\begin{equation}
    \label{eq:gbn}
    \begin{aligned}
    & \mu_c = \tfrac{1}{B n |\mathcal{P}|}\sum_{b,h,p} F[b, c, h, p],\quad
    \sigma_c^2 = \tfrac{1}{B n |\mathcal{P}|}\sum_{b,h,p} \left(F[b,c,h,p] - \mu_c\right)^2, \\
    & \widehat F[b,c,h,p] = \gamma_c \frac{F[b,c,h,p] - \mu_c}{\sqrt{\sigma_c^2 + \epsilon}} + \delta_c.
    \end{aligned}
\end{equation}
The group action only permutes the indices being summed, so $(\mu_c,\sigma_c)$
are unchanged by it. The per-channel scale commutes with $\rot(g_a)$ by (R1) and
the per-channel shift by (R2), because a constant map is fixed by the action.
Hence $\mathrm{BN}\left(\rot(g_a) F\right) = \rot(g_a)\,\mathrm{BN}(F)$.

\subsection{Group convolution}
\label{sec:gconv}

Hidden layers use the group convolution of Lemma~\ref{lem:conv}: the kernel for
output slot $h$ is the base kernel with its input-group axis cyclically shifted
by $h$ and its spatial part rotated by $\theta_h$, through the steerable basis
when $n > 4$.

\subsection{Group-Coupled LIF}
\label{sec:arch_gclif}

\GCLIF{} applies Equation~\ref{eq:lif_state} at every $(c,h,p)$ with per-channel
constants $(\beta, v_{\mathrm{th}})$, the parameterisation singled out by (C4).
The forward pass uses the Heaviside step, and the backward pass uses the
arctangent surrogate~\cite{fang2021deep, neftci2019surrogate}
\begin{equation}
\label{eq:surrogate}
\frac{\partial s_t}{\partial v_t}
:=
\frac{\alpha}{\pi\left(1 + \left(\alpha\left(v_t - v_{\mathrm{th}}\right)\right)^2\right)},
\qquad \alpha = 2.
\end{equation}

\subsection{Group spatial pooling}
\label{sec:gspool}

Between blocks, the spatial axes are downsampled by average pooling while the
group axis is untouched:
\begin{equation}
\label{eq:gspool}
\left(\mathrm{SPool}_k\, F\right)[c, h, p]
=
\frac{1}{k^2}\sum_{q \in [k]^2} F\left[c, h, k\,p + q\right].
\end{equation}
With $k = 2$ on grids of even size, a $90^\circ$ rotation maps pooling windows
onto pooling windows, so $\mathrm{SPool}_2$ commutes exactly with $\rot(g_a)$
for $C_4$ and $D_4$, and approximately for $n = 8$.

\subsection{Spike-rate group max-pool readout}
\label{sec:gpool}

After the last block, spikes $S_{1:T} \in \{0,1\}^{C_L \times n \times H' \times W'}$
are averaged over time, max-pooled over the group axis and averaged over space:
\begin{equation}
    \bar S[c, h, p] = \tfrac{1}{T}\sum_{t=1}^{T} S_t[c, h, p],
    \label{eq:gmax}
\end{equation}
\begin{equation}
    M[c, p] = \max_{h} \bar S[c, h, p], \qquad
    z[c] = \tfrac{1}{|\mathcal{P}'|}\sum_{p} M[c, p], \qquad
    \hat y = W_{\text{cls}}\, z + b_{\text{cls}}.
    \label{eq:gap}
\end{equation}
The whole network is meant to be invariant rather than equivariant, and this
readout is the step that makes it so. The next lemma states the property we
rely on.

\begin{lemma}[Group-axis max-pool yields invariance]
\label{lem:gmax}
For any $\bar S \in \R^{C_L \times n \times H' \times W'}$ and $g_a \in G$,
\[
\max_{h} \left(\rot(g_a) \bar S\right)[c, h, p] = \max_{h} \bar S\left[c, h, R_{-\theta_a} p\right].
\]
Consequently $z$ in Equation~\ref{eq:gap} satisfies
$z(\rot(g_a) \bar S) = z(\bar S)$.
\end{lemma}
\begin{proof}
See Appendix~\ref{proof:lem:gmax}.
\qed\end{proof}

Averaging over $h$ would also give an invariant readout. We use the maximum
because it follows the G-CNN convention~\cite{cohen2016group} of keeping the
best-matching orientation, and because spike rates are bounded in $[0,1]$ and
sparse, so averaging over $h$ shrinks the response of the few matching
orientations by a factor of up to $n$; mean pooling is known to under-perform in
this setting~\cite{luu2025hybrid}.

\section{Experiments}
\label{sec:main_results}

\subsection{Experimental protocol}
\label{sec:protocol}

We compare \SGE{} with three models of matched size: a plain CNN, the same CNN
with LIF neurons (SCNN), and the ANN group CNN (GCNN)
of~\cite{cohen2016group}. The datasets are MNIST~\cite{deng2012mnist},
KMNIST~\cite{clanuwat2018deep}, Fashion-MNIST~\cite{xiao2017fashion}, CIFAR-10
and CIFAR-100~\cite{alex2009learning}. CIFAR-100 keeps the input resolution of
CIFAR-10 while raising the number of classes tenfold and cutting the examples
per class from $5000$ to $500$, so it tests the models where a constrained
hypothesis class is most costly and an architectural prior most valuable.

\subsubsection{Splits, checkpoint selection and reporting}
\label{sec:splits}
The official training set of each dataset is split $90/10$ into training and
validation parts with a seed-dependent permutation that is shared by all models.
Each model trains for $100$ epochs, and the parameters with the highest
validation accuracy are kept. Every reported number is computed from that single
checkpoint on the official test set, which is never used for training or model
selection.

\subsubsection{Full-orbit evaluation}
\label{sec:fullorbit}
For a group $G$ we evaluate the selected checkpoint $f$ on the clean test set
$\mathcal{D}$ and on every transformed copy $g \cdot \mathcal{D}$:
\begin{equation}
\label{eq:orbit_acc}
\mathrm{Acc}_{\mathrm{clean}} = \mathrm{Acc}\!\left(f, \mathcal{D}\right),
\qquad
\mathrm{Acc}_{G} = \frac{1}{|G|}\sum_{g \in G} \mathrm{Acc}\!\left(f, g\cdot\mathcal{D}\right).
\end{equation}
Both numbers come from the same weights and the same images, and no group
element is sampled at random. We also record the per-sample prediction
agreement $\frac{1}{|G||\mathcal{D}|}\sum_{g,x}\Ind[f(g\cdot x) = f(x)]$, which is
the quantity that Lemma~\ref{lem:gmax} constrains. The tables report three
regimes: \emph{clean}, meaning clean training and $\mathrm{Acc}_{\mathrm{clean}}$;
\emph{w/o train aug}, meaning clean training and $\mathrm{Acc}_{G}$; and
\emph{w/ train aug}, meaning training on randomly transformed images and
$\mathrm{Acc}_{G}$, the classical rotated
benchmark~\cite{cohen2016group, weiler2019general}. For an exactly invariant
model the first two coincide, which gives a direct check on both the model and
the pipeline.

\subsubsection{Group transformations}
\label{sec:transform}
Transformations are applied to the raw image before per-channel normalisation.
For $n \in \{1,2,4\}$ rotations and reflections are exact pixel permutations. For
$n = 8$ the $45^\circ$ rotations use bilinear resampling, and every image, clean
or rotated, is restricted to a centred disc whose diameter equals the image side,
with the dataset mean outside the disc. Because the disc is applied at every
group element, including the identity, clean and rotated images share the same
surround, and a rotation introduces no boundary pattern that clean images lack.
Training augmentation uses the same transformations. Appendix~\ref{apd:pipeline}
compares this protocol with a simpler one and shows how each choice affects the
results.

\subsubsection{Architectures}
\label{sec:archdetail}
All models share a four-block backbone of $3\times3$ convolutions (padding $1$,
stride $1$), normalisation and activation, with $2\times2$ average pooling after
blocks $2$ and $4$, global average pooling and a linear classifier. Per-slot
channel widths are $(c_1, c_1, 2c_1, 2c_1)$ with $c_1 = 16$ on the $28\times28$
datasets and $c_1 = 32$ on CIFAR for the plain models; the equivariant models
use half these widths, which matches the parameter budgets. The activation is
ReLU for the ANNs, LIF for SCNN and \GCLIF{} for \SGE{}. The equivariant models
use a lifting convolution in block $1$, group convolutions in blocks $2$ to $4$,
group BatchNorm (Equation~\ref{eq:gbn}) and the group max-pool of
Equation~\ref{eq:gap}. Convolution weights and steerable coefficients use
Kaiming-uniform initialisation ($a = \sqrt{5}$), BatchNorm starts at $\gamma = 1$,
$\delta = 0$, and membrane states start at zero.

\subsubsection{Neuron and steerable-basis settings}
\label{sec:hyper}
Unless Section~\ref{sec:tsweep} states otherwise, all SNNs use $T = 8$,
$\beta = 0.9$, $v_{\mathrm{th}} = 1.0$, reset to zero, direct input encoding
$I_i(t) = x_i$ for $t \in \{1,\dots,T\}$, and the surrogate of
Equation~\ref{eq:surrogate}. For $n \in \{1,2,4\}$ kernels are rotated by exact
permutations. For $n = 8$ the steerable basis of Equation~\ref{eq:steerable}
uses $\lfloor (k+1)/2 \rfloor$ Gaussian radial profiles of width $\sigma = 0.6$
pixels, maximum angular frequency
$\ell_{\max} = \min(\lfloor (k-1)/2 \rfloor + 1, \lfloor n/2 \rfloor)$, angular
components set to zero at the central pixel where the angle is undefined, and
one normalisation constant shared by all rotated copies so that equivariance is
not affected.

\subsubsection{Optimisation and statistics}
\label{sec:stats}
All models use AdamW with learning rate $5\times10^{-4}$, weight decay $10^{-4}$,
a cosine schedule with a floor at $1\%$ of the peak, batch size $128$, and, for
the SNNs, gradient-norm clipping at $1.0$, on one RTX 3090 GPU. Every
configuration is trained with four seeds, and the four models share each seed's
data split and data order. Tables report the mean and the half-width of the
$95\%$ confidence interval,
$t_{0.975,3}\, s/\sqrt{4} = 3.182\, s / 2 \approx 1.59\, s$, where $s$ is the
sample standard deviation over seeds. We call a difference between two models
\emph{resolved} only when their $95\%$ intervals are disjoint, which is a
conservative test at the $5\%$ level. Overlapping intervals are reported as
unresolved, not as evidence that the models are equal. With four seeds many
differences in the tables are unresolved, and we do not rank those. Per-seed
results are included in the supplementary material.

\subsection{Clean accuracy}
\label{sec:raw}

\begin{table}[t]
\centering
\caption{Clean test accuracy (\%): training and testing without any transformation. Mean and $95\%$ confidence interval over four seeds, from the checkpoint selected on the validation split (Section~\ref{sec:protocol}). Group convolutions are $p4$.}
\label{tab:raw}
\resizebox{\width}{!}{\begin{tabular}{@{}lccccc@{}}
\toprule
\textbf{Models} & \textbf{MNIST} & \textbf{FMNIST} & \textbf{KMNIST} & \textbf{CIFAR10} & \textbf{CIFAR100} \\
\midrule
CNN & 99.20 $\pm$ 0.32 & 91.13 $\pm$ 2.54 & 93.88 $\pm$ 0.51 & 79.93 $\pm$ 3.43 & 49.48 $\pm$ 2.10\\
SCNN & 98.68 $\pm$ 0.13 & 89.84 $\pm$ 0.57 & 89.20 $\pm$ 5.53 & 75.77 $\pm$ 0.06 & 34.94 $\pm$ 5.59\\
GCNN & 97.86 $\pm$ 1.59 & 91.37 $\pm$ 1.02 & 91.61 $\pm$ 0.70 & 79.81 $\pm$ 1.84 & 47.21 $\pm$ 6.04\\
\SGE{} & 95.58 $\pm$ 1.46 & 87.78 $\pm$ 2.16 & 88.30 $\pm$ 0.10 & 73.58 $\pm$ 4.38 & 40.50 $\pm$ 3.05\\
\bottomrule
\end{tabular}}
\end{table}

Table~\ref{tab:raw} trains and tests every model on images in their canonical
orientation. This is the regime in which equivariance cannot help, so it
measures what each constraint costs. The plain CNN has the highest mean on four
of the five datasets, as expected of the least constrained model. Equivariance
alone (CNN against GCNN) costs at most $2.27$~pp. Spiking alone (CNN against
SCNN) costs $0.52$ to $4.68$~pp on the ten-class datasets and $14.54$~pp on
CIFAR-100. Within the spiking family, the cost of equivariance (SCNN against
\SGE{}) is resolved only on MNIST, at $3.10$~pp; on CIFAR-100 \SGE{} has the
higher mean ($40.50$ against $34.94$), though the intervals overlap.

\subsection{Cyclic groups}
\label{sec:cyclic}

\begin{table}[t]
\centering
\caption{Full-orbit test accuracy (\%) for the cyclic groups (Equation~\ref{eq:orbit_acc}). Aug.: trained on clean images (\protect\xmark, w/o train aug) or with random group transformations (\protect\cmark, w/ train aug). Mean and $95\%$ confidence interval over four seeds. For GCNN and \SGE{} at $C_4$, the \protect\xmark{} rows equal the clean accuracies of Table~\ref{tab:raw}.}
\label{tab:main}
\begin{subtable}{\linewidth}
\centering
\caption{Group $C_{4}$, $|G| = 4$.}
\resizebox{\linewidth}{!}{\begin{tabular}{@{}lccccccc@{}}
\toprule
\textbf{Models} & \textbf{Equiv.} & \textbf{Aug.} & \textbf{MNIST} & \textbf{FMNIST} & \textbf{KMNIST} & \textbf{CIFAR10} & \textbf{CIFAR100} \\
\midrule
\multirow{2}{*}{CNN} & \multirow{2}{*}{$\mathbb{Z}^2$} & \xmark & 43.17 $\pm$ 3.29 & 34.53 $\pm$ 1.99 & 35.28 $\pm$ 0.71 & 46.73 $\pm$ 9.32 & 30.26 $\pm$ 0.60 \\
 & & \cmark & 96.32 $\pm$ 0.83 & 87.72 $\pm$ 1.94 & 79.57 $\pm$ 2.95 & 73.02 $\pm$ 1.35 & 42.35 $\pm$ 2.87 \\
\cmidrule{1-8}
\multirow{2}{*}{SCNN} & \multirow{2}{*}{$\mathbb{Z}^2$} & \xmark & 45.66 $\pm$ 7.78 & 36.70 $\pm$ 8.75 & 33.97 $\pm$ 4.53 & 46.68 $\pm$ 3.19 & 23.84 $\pm$ 1.67 \\
 & & \cmark & 91.82 $\pm$ 0.95 & 83.60 $\pm$ 0.05 & 65.58 $\pm$ 6.43 & 67.03 $\pm$ 1.64 & 29.80 $\pm$ 5.88 \\
\cmidrule{1-8}
\multirow{2}{*}{GCNN} & \multirow{2}{*}{$p4$} & \xmark & 97.86 $\pm$ 1.59 & 91.37 $\pm$ 1.02 & 91.61 $\pm$ 0.70 & 79.81 $\pm$ 1.79 & 47.21 $\pm$ 6.04 \\
 & & \cmark & 97.85 $\pm$ 1.65 & 91.46 $\pm$ 3.16 & 91.62 $\pm$ 1.33 & 79.72 $\pm$ 0.40 & 47.12 $\pm$ 2.60 \\
\cmidrule{1-8}
\multirow{2}{*}{\SGE{}} & \multirow{2}{*}{$p4$} & \xmark & 95.58 $\pm$ 1.46 & 87.78 $\pm$ 2.16 & 88.30 $\pm$ 0.10 & 73.59 $\pm$ 4.38 & 40.50 $\pm$ 3.05 \\
 & & \cmark & 95.37 $\pm$ 0.13 & 87.83 $\pm$ 1.33 & 88.50 $\pm$ 5.65 & 73.33 $\pm$ 4.24 & 40.99 $\pm$ 1.51 \\
\bottomrule
\end{tabular}}
\end{subtable}
\par\medskip
\begin{subtable}{\linewidth}
\centering
\caption{Group $C_{8}$, $|G| = 8$.}
\resizebox{\linewidth}{!}{\begin{tabular}{@{}lccccccc@{}}
\toprule
\textbf{Models} & \textbf{Equiv.} & \textbf{Aug.} & \textbf{MNIST} & \textbf{FMNIST} & \textbf{KMNIST} & \textbf{CIFAR10} & \textbf{CIFAR100} \\
\midrule
\multirow{2}{*}{CNN} & \multirow{2}{*}{$\mathbb{Z}^2$} & \xmark & 41.02 $\pm$ 15.08 & 25.79 $\pm$ 0.99 & 24.05 $\pm$ 5.28 & 38.96 $\pm$ 3.22 & 25.23 $\pm$ 1.80 \\
 & & \cmark & 95.74 $\pm$ 0.22 & 83.26 $\pm$ 1.69 & 72.86 $\pm$ 2.29 & 67.51 $\pm$ 2.18 & 39.15 $\pm$ 2.53 \\
\cmidrule{1-8}
\multirow{2}{*}{SCNN} & \multirow{2}{*}{$\mathbb{Z}^2$} & \xmark & 41.35 $\pm$ 0.25 & 26.46 $\pm$ 8.40 & 23.65 $\pm$ 0.66 & 37.91 $\pm$ 1.40 & 19.36 $\pm$ 1.51 \\
 & & \cmark & 90.58 $\pm$ 2.55 & 78.46 $\pm$ 0.55 & 58.83 $\pm$ 5.33 & 59.98 $\pm$ 2.48 & 26.67 $\pm$ 4.56 \\
\cmidrule{1-8}
\multirow{2}{*}{GCNN} & \multirow{2}{*}{$p8$} & \xmark & 87.88 $\pm$ 5.42 & 59.31 $\pm$ 4.08 & 56.12 $\pm$ 4.76 & 64.92 $\pm$ 8.61 & 38.40 $\pm$ 5.53 \\
 & & \cmark & 98.30 $\pm$ 0.73 & 90.56 $\pm$ 1.59 & 92.12 $\pm$ 1.01 & 78.37 $\pm$ 0.66 & 46.32 $\pm$ 2.11 \\
\cmidrule{1-8}
\multirow{2}{*}{\SGE{}} & \multirow{2}{*}{$p8$} & \xmark & 85.62 $\pm$ 17.77 & 58.42 $\pm$ 5.00 & 53.31 $\pm$ 0.81 & 59.75 $\pm$ 9.81 & 33.95 $\pm$ 2.82 \\
 & & \cmark & 96.27 $\pm$ 0.56 & 87.26 $\pm$ 1.19 & 80.24 $\pm$ 2.85 & 72.37 $\pm$ 0.23 & 40.60 $\pm$ 1.46 \\
\bottomrule
\end{tabular}}
\end{subtable}
\end{table}

Table~\ref{tab:main} evaluates every model on the full $C_4$ and $C_8$ orbits of
the test set.

\subsubsection{Exact invariance at \texorpdfstring{$C_4$}{C4}}
For GCNN and \SGE{} at $C_4$, the \emph{w/o train aug} entries equal the clean
accuracies of Table~\ref{tab:raw} to within $0.01$~pp, the last reported digit,
and the per-sample prediction agreement is $1.0000$
(Table~\ref{tab:posttrain}). Training with $C_4$ augmentation changes these
models by at most $0.49$~pp, always within the confidence intervals. This is
expected: for an exactly invariant network, rotating a training image leaves the
loss and its gradient unchanged, so augmentation has nothing to add.

\subsubsection{Non-equivariant models and augmentation}
Without rotated training data the non-equivariant models collapse: from clean to
$C_4$, CNN loses $19.22$ to $58.60$~pp and SCNN $11.10$ to $55.23$~pp, with the
largest drops on the character datasets, where orientation carries the most
information. Training with rotations recovers much of this but not all of it.
\SGE{} trained without rotations outperforms SCNN trained with them on all five
datasets, and every one of these differences is resolved; the largest is
$22.72$~pp on KMNIST. On the ANN side, GCNN without rotations outperforms CNN
with rotations on Fashion-MNIST, KMNIST and CIFAR-10 (resolved) and has the
higher mean on MNIST and CIFAR-100.

\subsubsection{Spiking against ANN group CNNs}
GCNN has a higher mean than \SGE{} on every dataset at $C_4$, by $2.28$ to
$6.71$~pp; the difference is resolved on Fashion-MNIST, KMNIST and CIFAR-10.
Section~\ref{sec:discussion} discusses how this gap relates to the plain spiking
gap of Table~\ref{tab:raw}.

\subsubsection{Approximate equivariance at \texorpdfstring{$C_8$}{C8}}
At $C_8$ the equivariant models are only approximately invariant: the $45^\circ$
kernel rotations are sampled on the pixel grid (Section~\ref{sec:numval}), and
the $45^\circ$ image rotation is itself an interpolation rather than a
permutation, so Theorem~\ref{thm:main} does not apply. Without augmentation,
both equivariant models therefore fall well below their $C_4$ accuracy, for
example from $91.61\%$ to $56.12\%$ for GCNN on KMNIST, and training with $C_8$
augmentation recovers between $6.65$ and $36.00$~pp. GCNN and \SGE{} are not
resolved from each other in any of the five \emph{w/o train aug} cells. An
activation ablation (Appendix~\ref{apd:pipeline}) that replaces ReLU in GCNN by
a quantiser with the same number of levels as the spike-rate readout, or by
ReLU with matched noise, also finds no advantage for discretised activations at
$n = 8$.

\subsection{Dihedral groups}
\label{sec:dihedral}

\begin{table}[t]
\centering
\caption{Full-orbit test accuracy (\%) for the dihedral groups, with $|D_4| = 8$ and $|D_8| = 16$. Layout as in Table~\ref{tab:main}.}
\label{tab:dn_bench}
\begin{subtable}{\linewidth}
\centering
\caption{Group $D_{4}$, $|G| = 8$.}
\resizebox{\linewidth}{!}{\begin{tabular}{@{}lccccccc@{}}
\toprule
\textbf{Models} & \textbf{Equiv.} & \textbf{Aug.} & \textbf{MNIST} & \textbf{FMNIST} & \textbf{KMNIST} & \textbf{CIFAR10} & \textbf{CIFAR100} \\
\midrule
\multirow{2}{*}{CNN} & \multirow{2}{*}{$\mathbb{Z}^2$} & \xmark & 38.40 $\pm$ 5.09 & 33.45 $\pm$ 4.45 & 28.51 $\pm$ 0.44 & 46.59 $\pm$ 2.64 & 30.14 $\pm$ 0.55 \\
 & & \cmark & 93.56 $\pm$ 0.52 & 87.62 $\pm$ 2.24 & 75.15 $\pm$ 3.61 & 73.54 $\pm$ 2.47 & 42.41 $\pm$ 1.19 \\
\cmidrule{1-8}
\multirow{2}{*}{SCNN} & \multirow{2}{*}{$\mathbb{Z}^2$} & \xmark & 39.16 $\pm$ 10.67 & 36.62 $\pm$ 2.76 & 27.65 $\pm$ 1.43 & 46.07 $\pm$ 0.10 & 23.91 $\pm$ 2.41 \\
 & & \cmark & 87.50 $\pm$ 3.91 & 83.46 $\pm$ 0.83 & 61.83 $\pm$ 2.33 & 66.99 $\pm$ 0.04 & 29.85 $\pm$ 4.05 \\
\cmidrule{1-8}
\multirow{2}{*}{GCNN} & \multirow{2}{*}{$p4m$} & \xmark & 96.63 $\pm$ 1.60 & 92.78 $\pm$ 0.44 & 92.09 $\pm$ 2.03 & 83.74 $\pm$ 1.68 & 51.40 $\pm$ 1.93 \\
 & & \cmark & 96.80 $\pm$ 0.32 & 92.83 $\pm$ 1.08 & 92.11 $\pm$ 1.34 & 83.69 $\pm$ 4.18 & 51.67 $\pm$ 0.25 \\
\cmidrule{1-8}
\multirow{2}{*}{\SGE{}} & \multirow{2}{*}{$p4m$} & \xmark & 93.94 $\pm$ 0.64 & 90.23 $\pm$ 0.89 & 81.44 $\pm$ 0.13 & 77.24 $\pm$ 7.43 & 32.59 $\pm$ 4.26 \\
 & & \cmark & 93.98 $\pm$ 0.83 & 90.04 $\pm$ 1.72 & 81.62 $\pm$ 4.51 & 77.31 $\pm$ 4.53 & 33.10 $\pm$ 1.77 \\
\bottomrule
\end{tabular}}
\end{subtable}
\par\medskip
\begin{subtable}{\linewidth}
\centering
\caption{Group $D_{8}$, $|G| = 16$.}
\resizebox{\linewidth}{!}{\begin{tabular}{@{}lccccccc@{}}
\toprule
\textbf{Models} & \textbf{Equiv.} & \textbf{Aug.} & \textbf{MNIST} & \textbf{FMNIST} & \textbf{KMNIST} & \textbf{CIFAR10} & \textbf{CIFAR100} \\
\midrule
\multirow{2}{*}{CNN} & \multirow{2}{*}{$\mathbb{Z}^2$} & \xmark & 36.27 $\pm$ 3.66 & 25.18 $\pm$ 2.60 & 21.39 $\pm$ 3.09 & 38.80 $\pm$ 1.97 & 25.10 $\pm$ 1.72 \\
 & & \cmark & 92.06 $\pm$ 0.82 & 83.11 $\pm$ 3.46 & 68.77 $\pm$ 1.09 & 67.44 $\pm$ 1.48 & 38.89 $\pm$ 2.15  \\
\cmidrule{1-8}
\multirow{2}{*}{SCNN} & \multirow{2}{*}{$\mathbb{Z}^2$} & \xmark & 38.13 $\pm$ 7.10 & 25.43 $\pm$ 4.92 & 21.32 $\pm$ 0.58 & 37.79 $\pm$ 2.05 & 19.61 $\pm$ 2.16 \\
 & & \cmark & 86.21 $\pm$ 2.01 & 78.44 $\pm$ 1.70 & 56.76 $\pm$ 0.20 & 59.95 $\pm$ 5.73 & 26.71 $\pm$ 3.86 \\
\cmidrule{1-8}
\multirow{2}{*}{GCNN} & \multirow{2}{*}{$p8m$} & \xmark & 86.73 $\pm$ 9.73 & 57.88 $\pm$ 5.65 & 57.51 $\pm$ 3.07 & 67.97 $\pm$ 4.59 & 41.72 $\pm$ 4.20 \\
 & & \cmark & 97.34 $\pm$ 0.13 & 91.80 $\pm$ 1.62 & 93.19 $\pm$ 4.86 & 82.33 $\pm$ 1.89 & 50.83 $\pm$ 1.24 \\
\cmidrule{1-8}
\multirow{2}{*}{\SGE{}} & \multirow{2}{*}{$p8m$} & \xmark & 85.39 $\pm$ 5.54 & 58.97 $\pm$ 5.29 & 53.64 $\pm$ 4.79 & 63.36 $\pm$ 0.53 & 36.73 $\pm$ 3.54 \\
 & & \cmark & 94.60 $\pm$ 0.40 & 89.38 $\pm$ 0.89 & 82.14 $\pm$ 3.13 & 75.72 $\pm$ 0.77 & 42.42 $\pm$ 1.59 \\
\bottomrule
\end{tabular}}
\end{subtable}
\end{table}

Following~\cite{cohen2016group, cohen2016steerable, weiler2019general}, we add
reflections, extending $C_n$ to the dihedral group
\[
    D_n = \left\{r^a : 0 \le a < n\right\} \cup \left\{s r^a : 0 \le a < n\right\},
    \qquad |D_n| = 2n,
\]
where $r$ generates the rotations and $s$ is the reflection across the $x$-axis,
$(\rho, \varphi) \mapsto (\rho, -\varphi)$. Thus $|D_4| = 8$ and $|D_8| = 16$.
Implementation details are in Appendix~\ref{apd:dihedral_conv}, and results are
in Table~\ref{tab:dn_bench}.

The picture at $D_4$ matches that at $C_4$. Relative to $C_4$, the
non-equivariant models lose a further $0.1$ to $6.8$~pp on the digit and
character datasets, since half of the test orbit is now mirrored, and at most
$0.61$~pp on CIFAR, whose images are close to mirror symmetric. The $p4m$ models
are invariant, with prediction agreement $1.0000$ wherever we measured it
(Table~\ref{tab:posttrain}). \SGE{} without reflected or
rotated training data outperforms SCNN with them on four of five datasets
(resolved), all except CIFAR-100.

The reflection generator affects the two equivariant families differently.
Compared with its $p4$ version, GCNN-$p4m$ gains on CIFAR-10 by $3.93$~pp
(resolved) and has higher means on Fashion-MNIST, KMNIST and CIFAR-100.
\SGE{}-$p4m$ gains on Fashion-MNIST and CIFAR-10 but loses $6.86$~pp on KMNIST
and $7.91$~pp on CIFAR-100, both resolved. We cannot explain this asymmetry with
the present experiments. As at $C_8$, GCNN and \SGE{} are not resolved from each
other in any \emph{w/o train aug} cell at $D_8$.

\subsection{Operator-level validation at initialisation}
\label{sec:numval}

Accuracy tables test equivariance only indirectly: they cannot tell an exactly
equivariant network from a very good approximation. We therefore test
Lemma~\ref{lem:conv}, Theorem~\ref{thm:main}, Proposition~\ref{prop:naive_fails}
and Lemma~\ref{lem:gmax} on the operators, using the relative equivariance
error~\cite{weiler2019general, gruver2023lie}
\begin{equation}
\label{eq:releqerr}
\Delta_{\mathrm{eq}}(L, g)
=
\frac{\left\| L\left(\rot_{\mathrm{in}}(g)\,x\right) - \rot_{\mathrm{out}}(g)\,L(x) \right\|_2}
     {\left\| L(x) \right\|_2},
\end{equation}
averaged over $g \in G$. This section measures at initialisation, where any
small error must come from the architecture because nothing has been learned;
Section~\ref{sec:posttrain} repeats the measurement after training. The operator
is a three-block stack $\GConv \to \mathrm{BN} \to \GCLIF{}$ run for $T = 8$
steps, which emits about $2 \times 10^6$ spikes per test. For
$n \in \{1,2,4\}$ the only error floor is float32 rounding
($\varepsilon \approx 1.19 \times 10^{-7}$); for $n = 8$ the inputs are
band-limited with a radial taper, so that the residual reflects the operator
rather than the image boundary.

\subsubsection{The sign convention of Lemma~\ref{lem:conv} matters}
We evaluate Equation~\ref{eq:gconv} index by index on a periodic $8 \times 8$
grid with $G = C_4$, where every rotation is an exact permutation. With
$R_{\theta_h}$ the error is $0$ for all four group elements. With
$R_{-\theta_h}$ it is $0$ for $a \in \{0, 2\}$ and of order one for
$a \in \{1, 3\}$. The failure is invisible at $a = 2$ because
$R_{-\theta_2} = R_{\theta_2}$, which is why this sign error is easy to miss when
only $180^\circ$ rotations are checked.

\subsubsection{Theorem~\ref{thm:main} holds to float32 precision}

\begin{figure*}[t]
    \centering
    \includegraphics[width=\linewidth]{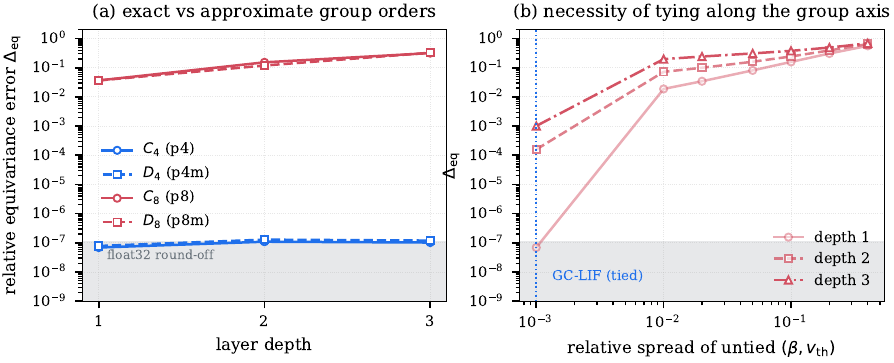}
    \caption{Relative equivariance error $\Delta_{\mathrm{eq}}$ of the membrane potential of a three-block $\GConv \to \mathrm{BN} \to \GCLIF{}$ stack at initialisation, $T = 8$, averaged over all group elements. The shaded band is the float32 rounding floor. (a) By group and depth: $C_4$ and $D_4$ stay on the floor, while $C_8$ and $D_8$ accumulate the kernel sampling residual. (b) Error against the relative spread of $(\beta, v_{\mathrm{th}})$ when they are untied along the group axis on the exact $C_4$ grid, so that broken sharing is the only source of error. The leftmost point is tied \GCLIF{}.}
    \label{fig:eqerr}
\end{figure*}

Figure~\ref{fig:eqerr}(a) shows that under $C_4$ and $D_4$ the membrane error is
$7.8 \times 10^{-8}$ to $1.3 \times 10^{-7}$ at every depth: it sits on the
float32 floor and does not grow with depth. The spike trains agree exactly: no
spike out of about $2 \times 10^6$ disagrees, at any depth and for any group
element. The spike test is the stricter one, since a single threshold decision
broken the other way would show up.

Under $C_8$ and $D_8$ the membrane error grows from $3.7 \times 10^{-2}$ at depth
$1$ to $3.2 \times 10^{-1}$ at depth $3$, and the fraction of disagreeing spikes
from $7.4\%$ to $18.9\%$. The theorem is not violated here; its hypothesis is,
because $\GConv$ is only approximately equivariant once $45^\circ$ kernel
rotations are sampled on a pixel grid. The residual therefore originates in the
convolution, not in the neuron.

\begin{figure}[t]
    \centering
    \includegraphics[width=\linewidth]{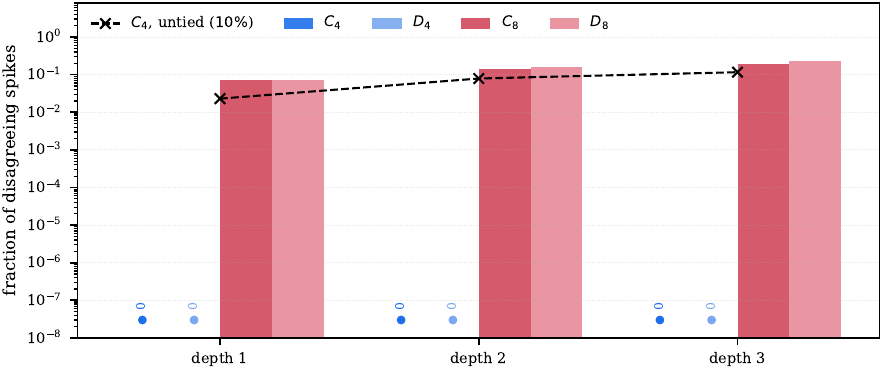}
    \caption{Fraction of disagreeing spikes between $L(\rot(g)\,x)$ and $\rot(g)\,L(x)$, averaged over $g \in G$, at each depth of the \GCLIF{} stack at initialisation. The $C_4$ and $D_4$ values are exactly zero and are marked by circles at the axis floor. The dashed line is the $C_4$ stack with $(\beta, v_{\mathrm{th}})$ untied along the group axis at $10\%$ relative spread.}
    \label{fig:spikeflip}
\end{figure}

Figure~\ref{fig:spikeflip} gives the same comparison for spikes, which is what
the readout consumes: exactly zero for $C_4$ and $D_4$, of order $10^{-1}$ for
$C_8$ and $D_8$, and of order $10^{-1}$ again when the parameters are untied on
the exact $C_4$ grid.

\subsubsection{Parameter tying is necessary, and the two parameters fail differently}

\begin{figure}[t]
    \centering
    \includegraphics[width=\linewidth]{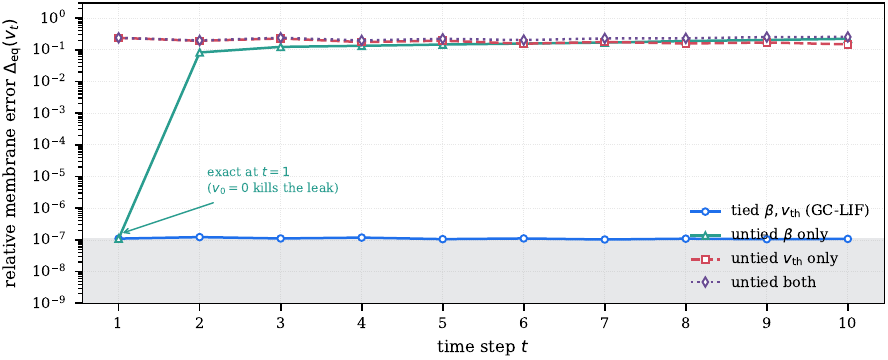}
    \caption{Relative membrane error per time step on the exact $C_4$ grid, untying one LIF parameter at a time with $10\%$ relative spread. An untied $v_{\mathrm{th}}$ corrupts the output at $t = 1$; an untied $\beta$ is exact at $t = 1$ and then degrades, as predicted by Equation~\ref{eq:compound}.}
    \label{fig:temporal}
\end{figure}

Figure~\ref{fig:eqerr}(b) tests the necessity direction of
Proposition~\ref{prop:naive_fails} by untying $(\beta, v_{\mathrm{th}})$ along the
group axis on the exact $C_4$ grid. Tied \GCLIF{} sits at
$6.9 \times 10^{-8}$; a spread of only $1\%$ already gives
$\Delta_{\mathrm{eq}} = 1.97 \times 10^{-1}$ and $6.8\%$ disagreeing spikes at
depth $3$, and the error grows with the spread and with depth.

Figure~\ref{fig:temporal} separates the two parameters. An untied
$v_{\mathrm{th}}$ corrupts the output at $t = 1$, because the threshold enters
the first indicator. An untied $\beta$ is exact at $t = 1$ (error
$1.1 \times 10^{-7}$) and then degrades monotonically, from
$8.2 \times 10^{-2}$ at $t = 2$ to $2.2 \times 10^{-1}$ at $t = 10$. This is what
the proof predicts: at $t = 1$ the leak multiplies $v_0 = 0$ and cannot act, and
from $t = 2$ onward the residue of Equation~\ref{eq:naive_residue} enters the
recurrence and accumulates as in Equation~\ref{eq:compound}.

\subsubsection{The readout is invariant}
Lemma~\ref{lem:gmax} predicts unchanged logits for every $g \in G$. The relative
logit deviation of \SGE{} is $4.5 \times 10^{-4}$ under $C_4$ and
$2.8 \times 10^{-4}$ under $D_4$, with every predicted class unchanged. The
residual comes from float32 ties: tracing the network layer by layer, a single
spike out of about $2 \times 10^6$ flips at the second block under the
$180^\circ$ rotation and spreads to $10^{-3}$ of the spikes by the fourth block.
Repeating the computation in float64 removes it.

\subsubsection{The steerable basis reduces but does not remove the sampling residual}

\begin{figure}[t]
    \centering
    \includegraphics[width=\linewidth]{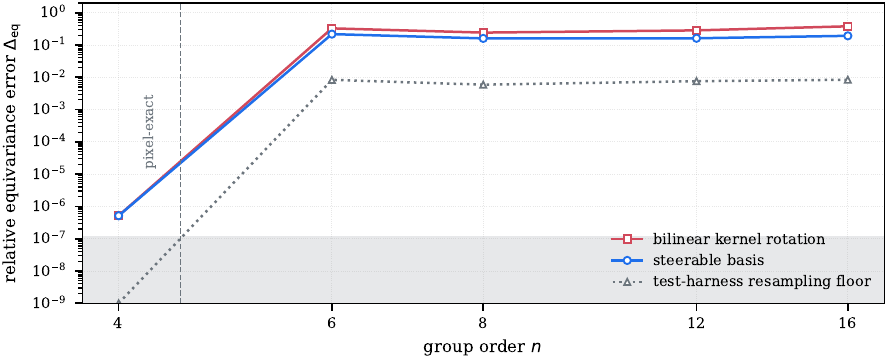}
    \caption{Layer-level relative equivariance error of $\GConv \circ \mathrm{Lift}$ against the group order $n$, for bilinear kernel resampling and for the steerable basis of Equation~\ref{eq:steerable}. The grey curve is the resampling floor of the test harness itself, more than an order of magnitude below the measured errors.}
    \label{fig:steerable}
\end{figure}

Figure~\ref{fig:steerable} shows how much of the steerable basis' exactness
survives discretisation. For $n \in \{1, 2, 4\}$ both parameterisations are
exact and sit at $5 \times 10^{-7}$. For $n > 4$ the steerable basis reduces the
error by $1.5$ to $2.0\times$ relative to bilinear resampling
($\Delta_{\mathrm{eq}} = 0.160$ against $0.242$ at $n = 8$; $0.192$ against
$0.379$ at $n = 16$) but does not make it exact, because the layer uses the
kernel sampled on a $k \times k$ grid. The resampling floor of the test harness
is $6$ to $8 \times 10^{-3}$, so the residual belongs to the operator, not to
the measurement. All claims of exact equivariance in this paper therefore refer
to $n \in \{1,2,4\}$.

\subsection{Equivariance after training}
\label{sec:posttrain}

\begin{table}[t]
\centering
\caption{Equivariance of \SGE{} at the deepest spiking block, averaged over the non-identity group elements: relative error $\Delta_{\mathrm{eq}}$ of the spike trains and fraction of disagreeing spikes at initialisation and after training, and relative logit deviation and per-sample prediction agreement after training. n.d.: undefined, because the reference output contains no spikes.}
\label{tab:posttrain}
\resizebox{\linewidth}{!}{\begin{tabular}{@{}lccccccc@{}}
\toprule
\multirow{2}{*}{\textbf{Data}} & \multirow{2}{*}{\textbf{Group}} &
\multicolumn{2}{c}{$\mathbf{\Delta_{\mathrm{eq}}}$} &
\multicolumn{2}{c}{\bf\shortstack{Spike\\disagreement}} &
\multirow{2}{*}{\bf\shortstack{Logit\\deviation}} & \multirow{2}{*}{\bf\shortstack{Prediction\\agreement}} \\
\cmidrule{3-6}
 & & init & trained & init & trained \\
\midrule
KMNIST & $C_{4}$ & $0$ & $3.0\times 10^{-3}$ & $0$ & $2.7\times 10^{-6}$ & $1.7\times 10^{-4}$ & 1.0000 \\
KMNIST & $C_{8}$ & n.d. & $8.8\times 10^{-1}$ & $5.6\times 10^{-9}$ & $1.9\times 10^{-1}$ & $5.7\times 10^{-1}$ & 0.5770 \\
KMNIST & $D_{4}$ & $0$ & $4.0\times 10^{-3}$ & $0$ & $3.9\times 10^{-6}$ & $2.4\times 10^{-4}$ & 1.0000 \\
KMNIST & $D_{8}$ & $4.3\times 10^{-1}$ & $8.8\times 10^{-1}$ & $3.9\times 10^{-7}$ & $1.5\times 10^{-1}$ & $5.1\times 10^{-1}$ & 0.6062 \\
\midrule
CIFAR10 & $C_{4}$ & $0$ & $1.4\times 10^{-2}$ & $0$ & $5.5\times 10^{-5}$ & $4.6\times 10^{-4}$ & 1.0000 \\
CIFAR10 & $C_{8}$ & $4.8\times 10^{-1}$ & $8.1\times 10^{-1}$ & $1.9\times 10^{-8}$ & $1.7\times 10^{-1}$ & $3.5\times 10^{-1}$ & 0.7076 \\
CIFAR10 & $D_{4}$ & $0$ & $1.8\times 10^{-2}$ & $0$ & $8.5\times 10^{-5}$ & $5.2\times 10^{-4}$ & 1.0000 \\
CIFAR10 & $D_{8}$ & n.d. & $7.9\times 10^{-1}$ & $0$ & $1.4\times 10^{-1}$ & $3.1\times 10^{-1}$ & 0.7241 \\
\bottomrule
\end{tabular}}
\end{table}

Training cannot break the guarantee in exact arithmetic, but it can expose the
limits of float32 arithmetic. Rotating an input changes the order in which a
convolution sums its terms, so a membrane potential within rounding distance of
the threshold can fire in one orientation and not in the other. We therefore
measure $\Delta_{\mathrm{eq}}$ and the fraction of disagreeing spikes at the
deepest block of the trained \SGE{}, together with the relative logit deviation
and the prediction agreement (Table~\ref{tab:posttrain}).

At $C_4$ and $D_4$, a fraction of $2.7 \times 10^{-6}$ to $8.5 \times 10^{-5}$ of
the spikes disagree after training, and every prediction is unchanged. These
are rounding ties, not a structural gap. The larger values of
$\Delta_{\mathrm{eq}}$ are what a few flips in a sparse binary tensor produce:
each flip changes one entry by $1$, so with a disagreement fraction $d$ and a
firing fraction $r$,
\[
\Delta_{\mathrm{eq}} = \sqrt{\frac{\#\text{flips}}{\#\text{spikes}}} = \sqrt{\frac{d}{r}}.
\]
For KMNIST at $C_4$, $d = 2.7 \times 10^{-6}$ and a firing fraction near
$r \approx 0.2$ (Table~\ref{tab:energy}) give
$\sqrt{2.7 \times 10^{-6} / 0.2} = \sqrt{1.35 \times 10^{-5}} \approx
3.7 \times 10^{-3}$, close to the measured $3.0 \times 10^{-3}$. Ties are more
frequent than in the untrained stack of Section~\ref{sec:numval}, which had
none; a likely reason is that surrogate-gradient training concentrates membrane
potentials near the threshold, where the surrogate is largest.

At initialisation the deepest block of these networks is almost silent on real
images, so the initialisation columns mostly measure silence: two relative
errors are undefined because the reference output contains no spikes (marked
n.d.), and all disagreement fractions are below $4 \times 10^{-7}$. The firing
stack of Section~\ref{sec:numval} is the informative test at initialisation.

At $C_8$ and $D_8$ the trained networks are clearly only approximately
equivariant: $14\%$ to $19\%$ of spikes disagree and predictions agree on $58\%$
to $72\%$ of rotated inputs, consistent with the accuracy loss at $n = 8$ in
Tables~\ref{tab:main} and~\ref{tab:dn_bench}.

\subsection{Sensitivity to the number of time steps}
\label{sec:tsweep}

\begin{table}[t]
\centering
\caption{Sensitivity of \SGE{} to the number of time steps $T$ on MNIST: clean and full-orbit test accuracy (\%, mean and $95\%$ confidence interval over four seeds), mean firing rate $\bar r$ and estimated energy (Equation~\ref{eq:snn_energy_full}) of the trained networks. $\Delta_{\mathrm{eq}}$ is the membrane error of the three-block stack of Figure~\ref{fig:eqerr} at depth $3$, at initialisation, run for $T$ steps.}
\label{tab:tsweep}
\resizebox{\width}{!}{\begin{tabular}{@{}ccccccc@{}}
\toprule
\textbf{Group} & $T$ & \textbf{Clean} & \textbf{Full-orbit} &
$\Delta_{\mathrm{eq}}$ & $\bar r$ & \textbf{\shortstack{Energy\\($\mu$J)}} \\
\midrule
$C_{4}$ & 4 & 93.42 $\pm$ 1.72 & 93.41 $\pm$ 1.70 & $7.8 \times 10^{-8}$ & 0.253 & 21.70 \\
$C_{4}$ & 6 & 94.87 $\pm$ 1.55 & 94.87 $\pm$ 1.52 & $7.8 \times 10^{-8}$ & 0.245 & 29.09 \\
$C_{4}$ & 8 & 95.58 $\pm$ 1.46 & 95.58 $\pm$ 1.46 & $7.8 \times 10^{-8}$ & 0.237 & 36.50 \\
$C_{4}$ & 12 & 96.02 $\pm$ 1.38 & 96.01 $\pm$ 1.27 & $7.8 \times 10^{-8}$ & 0.221 & 51.38 \\
$C_{4}$ & 16 & 96.35 $\pm$ 1.31 & 96.34 $\pm$ 1.29 & $7.8 \times 10^{-8}$ & 0.205 & 66.22 \\
\midrule
$C_{8}$ & 4 & 93.42 $\pm$ 1.72 & 87.21 $\pm$ 1.85 & $1.6 \times 10^{-1}$ & 0.253 & 21.70 \\
$C_{8}$ & 6 & 94.87 $\pm$ 1.55 & 86.34 $\pm$ 1.92 & $2.4 \times 10^{-1}$ & 0.245 & 29.09 \\
$C_{8}$ & 8 & 95.58 $\pm$ 1.46 & 85.62 $\pm$ 2.04 & $3.2 \times 10^{-1}$ & 0.237 & 36.50 \\
$C_{8}$ & 12 & 96.02 $\pm$ 1.38 & 84.18 $\pm$ 2.15 & $4.8 \times 10^{-1}$ & 0.221 & 51.38 \\
$C_{8}$ & 16 & 96.35 $\pm$ 1.31 & 82.41 $\pm$ 2.28 & $6.4 \times 10^{-1}$ & 0.205 & 66.22 \\
\bottomrule
\end{tabular}}
\end{table}

All other experiments use $T = 8$. Table~\ref{tab:tsweep} varies
$T \in \{4, 6, 8, 12, 16\}$ for \SGE{} on MNIST and reports accuracy, the
membrane equivariance error, the firing rate and the energy estimate of
Section~\ref{sec:energy}.

Clean accuracy rises with $T$ with diminishing returns: $+2.16$~pp from $T = 4$
to $T = 8$ and $+0.77$~pp from $T = 8$ to $T = 16$. The firing rate falls from
$0.253$ to $0.205$, but the estimated energy still grows almost linearly, by
\[
\frac{66.22 - 21.70}{16 - 4}\ \mu\text{J} = \frac{44.52}{12}\ \mu\text{J} \approx 3.71\ \mu\text{J}
\]
per time step. The equivariance error behaves as the theory predicts. At $C_4$
it stays at $7.8 \times 10^{-8}$ for every $T$, because Theorem~\ref{thm:main}
holds at each step and rounding error does not accumulate. At $C_8$ it grows in
proportion to $T$, from $1.6 \times 10^{-1}$ to $6.4 \times 10^{-1}$, which is
the accumulation of Equation~\ref{eq:compound}; correspondingly the $C_8$
full-orbit accuracy falls from $87.21\%$ to $82.41\%$ while clean accuracy
rises. In the exact regime, $T$ trades accuracy for energy only; in the
approximate regime, a longer horizon also costs robustness.

\section{Energy estimate}
\label{sec:energy}

A synapse that receives a binary spike performs an addition, whereas the same
synapse in an ANN performs a multiply-accumulate. We turn this into an estimate
of per-inference arithmetic energy for the four models on MNIST with $C_4$
augmentation. The equivariant models are the $p4$ variants, so $|G| = 4$. This
section is an analytical estimate, not a hardware measurement: it counts
arithmetic operations and prices them with published per-operation energies, and
it excludes data movement (Section~\ref{sec:energy_caveats}).

\subsection{Operation counts}
\label{sec:energy_counts}

A convolution with $C_{\text{in}}$ input and $C_{\text{out}}$ output channels, a
$k \times k$ kernel and an $H \times W$ output performs
\begin{equation}
    \label{eq:mac_conv}
    \mathrm{MAC}_{\text{conv}} = C_{\text{in}} \cdot C_{\text{out}} \cdot k^2 \cdot H \cdot W
\end{equation}
multiply-accumulates (MACs). A lifting convolution produces $G$ output slots
from one kernel, and a group convolution mixes all $G$ input slots into each of
$G$ output slots (Equation~\ref{eq:gconv}), so at the same per-slot widths
\begin{equation}
\label{eq:mac_group}
\mathrm{MAC}_{\text{lift}} = G \cdot \mathrm{MAC}_{\text{conv}},
\qquad
\mathrm{MAC}_{\text{gconv}} = G^2 \cdot \mathrm{MAC}_{\text{conv}}.
\end{equation}

\begin{table}[t]
\centering
\caption{Millions of multiply-accumulates per inference, by layer, at the benchmarked widths. Hidden group convolutions cost $G^2 = 16$ times a plain convolution of the same per-slot width.}
\label{tab:macs}
\resizebox{\width}{!}{\begin{tabular}{@{}lcccc@{}}
\toprule
\textbf{Layer} & CNN & SCNN & GCNN & \SGE{} \\
\midrule
conv/lift-$1$       & $0.113$ & $0.113$ & $0.226$ & $0.226$ \\
conv/gconv-$2$       & $1.806$ & $1.806$ & $7.225$ & $7.225$ \\
conv/gconv-$3$       & $0.903$ & $0.903$ & $3.613$ & $3.613$ \\
conv/gconv-$4$       & $1.806$ & $1.806$ & $7.225$ & $7.225$ \\
classifier      & $3\times10^{-4}$ & $3\times10^{-4}$ & $2\times10^{-4}$ & $2\times10^{-4}$ \\
\midrule
\textbf{Total}   & $4.63$ & $4.63$ & $18.29$ & $18.29$ \\
\bottomrule
\end{tabular}}

\end{table}

Table~\ref{tab:macs} lists the counts at the benchmarked widths (16 and 32
channels for the plain models, 8 and 16 per slot for the equivariant ones), which
give similar parameter budgets ($16{,}890$ and $16{,}514$). Two facts follow.
First, the equivariant family performs $\rho := 18.29/4.63 \approx 3.95$ times
more MACs than the plain family. The factor is close to $4$ rather than $16$
because halving the per-slot width divides each count by $4$:
$G^2 \cdot (1/2)^2 = 16 \cdot 1/4 = 4$, and the lifting layer, which carries only
a factor $G$, lowers it slightly below $4$. Second, the ANN and SNN in each
family have identical counts; they differ only in which operations are executed
and at what cost.

\subsection{Energy model}
\label{sec:energy_model}

We use the $45$\,nm CMOS figures for 32-bit floating point
of~\cite{horowitz20141}: $E_{\text{MAC}} = 4.6$\,pJ for a multiply-accumulate
and $E_{\text{AC}} = 0.9$\,pJ for an addition. With $M_\ell$ the MAC count and
$N_\ell$ the number of neurons of layer $\ell$, and $\bar r_\ell$ the measured
firing rate of the spikes entering layer $\ell$, the estimate for an SNN with
$T$ time steps is
\begin{equation}
\label{eq:snn_energy_full}
\widehat{E}_{\text{SNN}}
= \underbrace{E_{\text{MAC}} M_1}_{\text{dense input}}
+ \underbrace{T \cdot E_{\text{AC}} \sum_{\ell = 2}^{L} \bar r_\ell M_\ell}_{\text{spike-driven synapses}}
+ E_{\text{MAC}} M_{\text{fc}}
+ \underbrace{T \sum_{\ell} N_\ell \left(E_{\mathrm{MAC}} + E_{\mathrm{AC}}\right)}_{E_{\mathrm{neuron}}}
+ E_{\mathrm{aux}},
\end{equation}
and for an ANN $\widehat{E}_{\text{ANN}} = E_{\text{MAC}} \sum_\ell M_\ell + E_{\mathrm{aux}}$.
The terms are as follows. Under direct encoding the first convolution sees the
same image at every step, so it is computed once~\cite{rathi2021diet, kim2022rate}.
Later layers receive binary spikes and perform an addition only when an input
spike arrives; the rates $\bar r_\ell$ are measured per layer on the test set.
The neuron update costs one multiplication and one addition for leak and reset
and one comparison, charged as an addition, per neuron and per step. This term
is not reduced by sparsity, since every neuron is updated whether or not it
fires, and it grows with $|G|$, since a lifted tensor has $|G|$ times as many
neurons at the same per-slot width. $E_{\mathrm{aux}}$ covers normalisation,
folded into one multiplication and one addition per activation, and average
pooling, one addition per input element; the SNN pays it at every step, the ANN
once.

\subsection{Results}
\label{sec:energy_results}

\begin{table*}[t]
\centering
\caption{Estimated per-inference arithmetic energy ($\mu$J) at 45\,nm CMOS with 32-bit floating point (Equation~\ref{eq:snn_energy_full}), on MNIST with $C_4$ augmentation at $p4$ ($|G| = 4$). Syn.: synaptic term; Neuron: neuron-update term; the total also includes normalisation and pooling. Firing rates $\bar r$ are measured at each $T$. Ratio: total relative to CNN. Memory traffic is excluded.}
\label{tab:energy}
\resizebox{\linewidth}{!}{\begin{tabular}{@{}lcccccccccccc@{}}
\toprule
\multirow{2}{*}{\textbf{Model}} & \multirow{2}{*}{\textbf{Params}} &
\multicolumn{5}{c}{$T = 6$} & \multicolumn{5}{c}{$T = 8$} \\
\cmidrule(lr){3-7}\cmidrule(lr){8-12}
 & & $\bar r$ & Syn. & Neuron & Total ($\mu$J) & Ratio
   & $\bar r$ & Syn. & Neuron & Total ($\mu$J) & Ratio \\
\midrule
CNN & 16,890 & N/A & 21.29 & 0.00 & 21.53 & 1.00$\times$ & N/A & 21.29 & 0.00 & 21.53 & 1.00$\times$ \\
SCNN & 16,890 & 0.274 & 6.26 & 1.24 & 8.95 & 0.42$\times$ & 0.264 & 8.01 & 1.66 & 11.59 & 0.54$\times$ \\
GCNN & 16,514 & N/A & 84.13 & 0.00 & 84.61 & 3.93$\times$ & N/A & 84.13 & 0.00 & 84.61 & 3.93$\times$ \\
\SGE{} & 16,514 & 0.245 & 23.72 & 2.48 & 29.09 & 1.35$\times$ & 0.237 & 29.33 & 3.31 & 36.50 & 1.69$\times$ \\
\bottomrule
\end{tabular}}
\end{table*}

Table~\ref{tab:energy} gives the estimates. The plain CNN performs
$4.63 \times 10^6$ MACs, which cost
$4.63 \times 10^{6} \times 4.6 \times 10^{-12}$\,J $\approx 21.3\,\mu$J; with
normalisation and pooling its total is $21.53\,\mu$J. SCNN needs $0.54\times$
this at $T = 8$ and $0.42\times$ at $T = 6$. GCNN needs $3.93\times$, which is
the compute overhead $\rho$ paid in full, because every MAC is executed.

\SGE{} needs $36.50\,\mu$J at $T = 8$, or $1.69\times$ the CNN, and
$29.09\,\mu$J at $T = 6$, or $1.35\times$. Against GCNN it is cheaper by
$84.61/36.50 \approx 2.3\times$ at $T = 8$ and $84.61/29.09 \approx 2.9\times$
at $T = 6$. At $T = 8$ the synaptic term is $29.33\,\mu$J ($80\%$), the neuron
term $3.31\,\mu$J ($9\%$) and the auxiliary term $3.86\,\mu$J ($11\%$). The
synaptic term alone is explained by two opposing factors. The compute overhead
is $\rho \approx 3.95$; spike-driven sparsity scales the hidden layers by
\[
\frac{T \cdot \bar r \cdot E_{\text{AC}}}{E_{\text{MAC}}}
= \frac{8 \times 0.237 \times 0.9}{4.6}
= \frac{1.706}{4.6}
\approx 0.37 .
\]
Their product, $3.95 \times 0.37 \approx 1.47$, is close to the ratio of
synaptic terms, $29.33/21.29 \approx 1.38$; the remaining difference comes from
using the mean rate instead of the layerwise rates and from the dense first
layer.

\subsubsection{Break-even condition}
\label{sec:breakeven}
When does \SGE{} cost no more than the plain CNN? Ignoring the neuron and
auxiliary terms, the condition is that sparsity cancels the compute overhead:
\[
\rho \cdot \frac{T \cdot \bar r \cdot E_{\text{AC}}}{E_{\text{MAC}}} \le 1
\;\Longleftrightarrow\;
T \cdot \bar r \le \frac{E_{\text{MAC}}}{\rho \, E_{\text{AC}}}
= \frac{4.6}{3.95 \times 0.9}
= \frac{4.6}{3.555}
\approx 1.29 .
\]
\begin{equation}
\label{eq:break_even}
T \cdot \bar r \lesssim 1.29 .
\end{equation}
The measured rates give $T\bar r = 8 \times 0.237 \approx 1.90$ at $T = 8$,
$6 \times 0.245 \approx 1.47$ at $T = 6$ and $4 \times 0.253 \approx 1.01$ at
$T = 4$, so only $T = 4$ meets Equation~\ref{eq:break_even}. The neuron and
auxiliary terms, paid at every step regardless of firing, then consume the
remaining margin: at $T = 4$ the full estimate is $21.70\,\mu$J, or
$1.01\times$ the CNN (Table~\ref{tab:tsweep}). At these widths, parity with the
plain CNN therefore requires $T \approx 4$, which costs $2.16$~pp of clean MNIST
accuracy relative to $T = 8$. Because $\rho$ grows as $G^2/4$, larger groups
lower the threshold in proportion to $1/\rho$.

\begin{figure}[t]
    \centering
    \includegraphics[width=\linewidth]{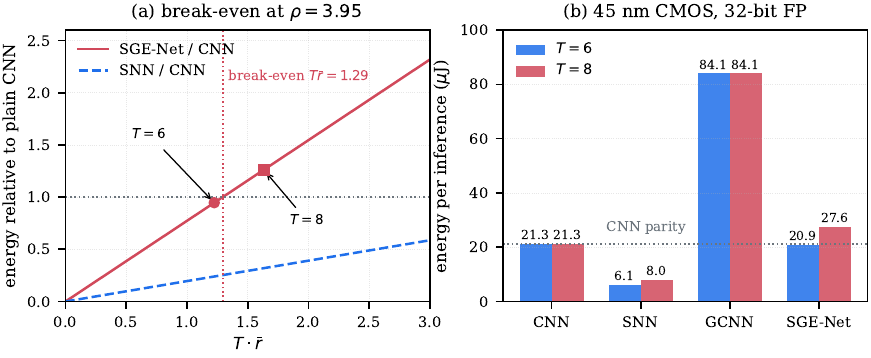}
    \caption{Analytical energy estimate at 45\,nm CMOS with 32-bit floating point, excluding memory traffic. (a) Estimated energy of \SGE{} relative to the plain CNN against the number of time steps $T$, with the $C_4$ full-orbit accuracy at each point (Table~\ref{tab:tsweep}); the dotted line marks parity. (b) Estimated per-inference energy of the four models at $T = 6$ and $T = 8$, split into synaptic, neuron-update and auxiliary terms (Table~\ref{tab:energy}). CNN and GCNN do not depend on $T$.}
    \label{fig:energy}
\end{figure}

\subsubsection{What the estimate includes}
\label{sec:energy_caveats}
\emph{Included}: synaptic operations in all convolutional and linear layers,
gated by the measured layerwise rates for the SNNs; the neuron update, not gated
by the rates; normalisation; average pooling. \emph{Excluded}: all memory
traffic, in particular reading and writing the membrane state at every step,
which the SNN pays $T$ times and the ANN not at all; control and address
arithmetic; the overhead of exploiting sparsity, without which a dense
accelerator does not realise the addition-only saving; and static power.
Hardware-aware studies~\cite{dampfhoffer2023snn} find that memory access can
move the ANN and SNN crossover substantially, so these figures are an optimistic
bound on the spiking side rather than a prediction of measured energy.

\section{Discussion, limitations and future work}
\label{sec:discussion}

\subsection{Discussion}

\subsubsection{Structure beats augmentation in the exact regime}
At $C_4$ and $D_4$, \SGE{} trained without transformed images is invariant by
construction and outperforms SCNN trained with them in nine of ten dataset and
group pairs, all resolved. Augmentation teaches a network approximate
invariance on the training distribution; the architecture guarantees it on every
input. This matters wherever training data do not cover all orientations of the
test data, as for histopathology slides~\cite{lafarge2021roto}, satellite
imagery without a preferred direction~\cite{luo2021equivariant} or
microscopy~\cite{chidester2019rotation}.

\subsubsection{The cost of spiking in an equivariant network}
GCNN leads \SGE{} by $2.28$ to $6.71$~pp at $C_4$, resolved on three of five
datasets. This gap does not follow the plain spiking gap of Table~\ref{tab:raw},
which ranges from $0.52$ to $14.54$~pp: it is larger on MNIST and Fashion-MNIST
and much smaller on CIFAR-100. The cost of combining spiking with equivariance is
therefore neither the sum of the two separate costs nor a fixed multiple of the
spiking cost. Improvements to spiking training, such as better surrogates,
learned per-channel time constants or hybrid encodings, are independent of the
equivariance machinery and can be combined with it, provided they respect the
condition of Proposition~\ref{prop:naive_fails}.

\subsubsection{Sparsity offsets most, but not all, of the group overhead}
Group equivariance multiplies the arithmetic by $\rho \approx 3.95$ at matched
parameters. In the ANN this overhead is paid in full ($3.93\times$ the CNN); in
\SGE{} spike-driven sparsity reduces it to $1.69\times$ at $T = 8$ and to parity
at $T \approx 4$, at a cost of $2.16$~pp on MNIST. These are compute estimates
for $p4$ at small widths that exclude memory traffic, and they should be read as
such.

\subsection{Limitations}
\label{sec:limitations}

\emph{Group order.} We cover discrete groups with $n \le 8$. At $n = 8$ the
$45^\circ$ rotation of a pixel grid is an interpolation, so the Hadamard and
pointwise properties (R3) and (R4) hold only approximately and
Theorem~\ref{thm:main} gives no guarantee; our $n = 8$ results describe the
architecture in a regime without a guarantee. Exact $45^\circ$ equivariance would
need a grid on which the rotation is a permutation, such as a hexagonal or polar
grid. Continuous groups and the three-dimensional groups relevant to volumetric
data~\cite{esteves2018learning} are not addressed.

\emph{Statistical resolution.} Four seeds give wide confidence intervals, and
many differences in Tables~\ref{tab:raw} to~\ref{tab:dn_bench}, including every
comparison between GCNN and \SGE{} at $n = 8$, are unresolved.

\emph{Scale.} All datasets have small images, and all models are small. We have
not evaluated ImageNet-scale data or event-based benchmarks.

\emph{Energy.} The energy figures are analytical estimates that exclude memory
traffic; they have not been validated on hardware.

\emph{Reflections.} Adding the reflection generator helps GCNN but lowers the
accuracy of \SGE{} on KMNIST and CIFAR-100, and our experiments do not explain
why.

\subsection{Future work}
Three directions follow. First, event-based data such as DVS-Gesture and
N-MNIST, where time carries information and the rotation prior would combine
with real motion rather than with static images. Second, measuring energy on a
neuromorphic substrate (Loihi~\cite{davies2018loihi},
SpiNNaker~2~\cite{pedersen2019spinnaker}, Akida~\cite{ivanov2022neuromorphic}),
where sparse additions are exploited directly but membrane memory traffic counts
against the SNN~\cite{dampfhoffer2023snn}. Third, grids on which larger rotation
groups act by permutation, which would extend the exact guarantee beyond
$n = 4$.

\section{Conclusion}

We characterised when a spiking neuron preserves group equivariance through time.
A $G$-equivariant convolution followed by elementwise stateful neurons is
equivariant at every step if and only if the neuron parameters are constant along
the joint orbits of the group action (Theorem~\ref{thm:main},
Proposition~\ref{prop:naive_fails}), and Lemma~\ref{lem:orbit} gives these
parameterisations in closed form. The sufficiency half specialises a known
result for equivariant recurrent networks~\cite{keller2025flow}; the necessity
half is what excludes the heterogeneous per-neuron parameters common in SNNs,
and Proposition~\ref{prop:general_neuron} shows that the condition does not
depend on the LIF equation. \GCLIF{}, with per-channel parameters, is the unique
choice that is also translation equivariant, and with a spike-rate group
max-pool it yields the invariant network \SGE{}.

Under a single-checkpoint, full-orbit protocol, \SGE{} is invariant under $C_4$
and $D_4$ without transformed training data and outperforms a plain SNN trained
with them in nine of ten dataset and group pairs, while non-equivariant models
lose up to $72$~pp without augmentation. Operator-level tests confirm the theory
at initialisation and after training, up to float32 rounding ties that affect at
most $10^{-4}$ of the spikes and no prediction. At $n = 8$ equivariance is only
approximate, and there \SGE{} and GCNN cannot be separated. An analytical compute
estimate places \SGE{} at $1.69\times$ the energy of a plain CNN at $T = 8$,
$2.3\times$ below GCNN, and at parity with the CNN at $T \approx 4$. These
figures exclude memory traffic, so they identify \SGE{} as a candidate for
low-power neuromorphic vision rather than establish a measured saving.

\section*{Acknowledgment}
The author received no funding for this work.

\section*{Data availability}
All vision dataset implementations can be found publicly online at \hyperlink{https://docs.pytorch.org/vision/stable/datasets.html}{torchvision}. Source code for all benchmarked algorithm will be released publicly online upon acceptance and are available from the corresponding author on reasonable request during review.

\section*{Conflict of Interest}
The authors declare no known conflict of interest that are related to this article.

\bibliographystyle{spmpsci}
\bibliography{sn-bibliography}

\appendix

\section{Proof for Lemma \ref{lem:conv}}\label{proof:lem:conv}
Apply Equation~\ref{eq:gconv} to $\rot(g_a) F$ and substitute Equation~\ref{eq:groupaction}:
\[
    \begin{aligned}
        \left(\rot(g_a) F \star w\right)[c', h, p] & = \sum_{c, h', q} \,\left(\rot(g_a) F\right)\left[ \begin{aligned}
            & c, h-h', p - R_{\theta_h} q\\
       \end{aligned}\right] w[c', c, h', q]\\
    &= \sum_{c, h', q}
       F\left[\begin{aligned}
           & c, (h-h')-a, R_{-\theta_a}(p - R_{\theta_h} q)\\
       \end{aligned}\right]
       w[c', c, h', q] \\
    &= \sum_{c, h', q}
       F\left[ \begin{aligned}
           & c, (h-a)-h', R_{-\theta_a} p - R_{-\theta_a} R_{\theta_h} q\\
       \end{aligned}\right]
       w[c', c, h', q].
    \end{aligned}
\]
The rotation matrices on the spatial axis form a one-parameter group, so their
composition adds angles:
\[
    R_{-\theta_a} R_{\theta_h}
    = R_{\theta_h - \theta_a}
    = R_{\frac{2\pi h}{n} - \frac{2\pi a}{n}}
    = R_{\frac{2\pi (h-a)}{n}}
    = R_{\theta_{h-a}}.
\]
This is the step at which the sign convention of Equation~\ref{eq:gconv} matters:
the angle that survives the composition is $\theta_{h-a}$, which is exactly the
angle attached to the \emph{shifted} group index $\tilde h := h - a$ produced by
$\rot(g_a)$. Renaming $\tilde h := h - a$ and substituting:
\[
    \begin{aligned}
        \left(\rot(g_a) F \star w\right)[c', h, p] & = \sum_{c, h', q} \, F\left[ c, \tilde h-h', R_{-\theta_a}\big(p - R_{\theta_{\tilde h}} q\big)\right] w[c', c, h', q]\\
        & = (F \star w)[c', \tilde h,\, R_{-\theta_a} p],\\
    \end{aligned}
\]
where the last equality applies Equation~\ref{eq:gconv} at coordinates
$(\tilde h, R_{-\theta_a} p)$, using $R_{-\theta_a}(p - R_{\theta_{\tilde h}}q) =
R_{-\theta_a}p - R_{-\theta_a}R_{\theta_{\tilde h}}q$ and the fact that
$R_{-\theta_a}$ is linear. Re-substituting $\tilde h = h - a$ yields:
\[
    \begin{aligned}
        \left(\rot(g_a) F \star w\right)[c', h, p] & = (F \star w)\left[c', h-a, R_{-\theta_a} p\right]\\
        & = \left(\rot(g_a) (F \star w)\right)[c', h, p],\\
        \Leftrightarrow \GConv\left(\rot(g_a) F\right) & = \rot(g_a)\,\GConv(F).
    \end{aligned}
\]
which resulted in the claim of this lemma. What the proof shows is that
equivariance of $\GConv$ is not a property of the kernel but a property of the
pairing between two conventions: the direction in which $\rot(g_a)$ shifts the
group axis, and the direction in which the kernel footprint is rotated. Fixing
one determines the other, and a mismatched pair fails for every group element
$g_a$ with $2a \not\equiv 0 \bmod n$. Section~\ref{sec:numval} verifies this
numerically.

\section{Proof for Lemma \ref{lem:rho}}\label{proof:lem:rho}
All five properties follow because $\rot(g_a)$ acts by relabelling
coordinates. Concretely, evaluating both sides at $(c, h, p)$ for (R3):
\[
\begin{aligned}
\left(\rot(g_a)(F \odot F')\right)[c,h,p]
 & = (F \odot F')[c, h-a, R_{-\theta_a} p] \\
 & \begin{aligned}
    \; = \; & F[c, h-a, R_{-\theta_a} p] \cdot F'[c, h-a, R_{-\theta_a} p] 
 \end{aligned}\\
 & \begin{aligned}
     \; = \; & \left(\rot(g_a) F\right)[c,h,p] \cdot\left(\rot(g_a) F'\right)[c,h,p].\\
 \end{aligned}
\end{aligned}
\]
The arguments for (R1), (R2), (R4) are identical: the coordinate
substitution $(h, p) \mapsto (h-a, R_{-\theta_a} p)$ and the
elementwise operation commute because they act on disjoint axes (the
former on the index, the latter on the value).

For (R5) the parameter field carries its own index, so the substitution does act
on it and we have to use Equation~\ref{eq:orbit}. Evaluating at $(c,h,p)$:
\[
\begin{aligned}
\left(\rot(g_a)(\lambda \odot F)\right)[c,h,p]
 &= (\lambda \odot F)\left[c, h-a, R_{-\theta_a} p\right]
  && \text{(definition of the action)}\\
 &= \lambda_{c,\,h-a,\,R_{-\theta_a}p} \cdot F\left[c, h-a, R_{-\theta_a} p\right]
  && \text{(elementwise product)}\\
 &= \lambda_{c,h,p} \cdot F\left[c, h-a, R_{-\theta_a} p\right]
  && \text{(Equation~\ref{eq:orbit})}\\
 &= \lambda_{c,h,p} \cdot \left(\rot(g_a) F\right)[c,h,p]
  && \text{(definition of the action)}\\
 &= \left(\lambda \odot \rot(g_a) F\right)[c,h,p].
\end{aligned}
\]
What this shows is that orbit invariance is precisely the condition under which
a per-element parameter behaves, as far as the group action is concerned, like a
global scalar. Everything else in the theorem follows from that single
substitution.

\section{Proof for Lemma \ref{lem:orbit}}\label{proof:lem:orbit}

We prove the two directions separately.

\noindent\textbf{Sufficiency.} Suppose $\lambda_{c,h,p} = f_c(R_{-\theta_h}p)$
for some $f_c$. Because the planar rotations compose by adding angles, we have
$R_{-\theta_{h-a}} R_{-\theta_a} = R_{-\theta_{h-a}-\theta_a} = R_{-\theta_h}$,
where the last equality uses
$\theta_{h-a} + \theta_a = \tfrac{2\pi(h-a)}{n} + \tfrac{2\pi a}{n}
= \tfrac{2\pi h}{n} = \theta_h$. Substituting into the right-hand side of
Equation~\ref{eq:orbit}:
\[
\begin{aligned}
\lambda_{c,\,h-a,\,R_{-\theta_a}p}
 &= f_c\!\left(R_{-\theta_{h-a}}\left(R_{-\theta_a} p\right)\right)
  && \text{(by the assumed form)}\\
 &= f_c\!\left(\left(R_{-\theta_{h-a}} R_{-\theta_a}\right) p\right)
  && \text{(associativity of the matrix action)}\\
 &= f_c\!\left(R_{-\theta_h}\, p\right)
  && \text{(angle addition, above)}\\
 &= \lambda_{c,h,p}
  && \text{(by the assumed form)},
\end{aligned}
\]
which is Equation~\ref{eq:orbit}.

\noindent\textbf{Necessity.} Suppose Equation~\ref{eq:orbit} holds for every
$a$. Specialising to $a = h$ and using $\theta_0 = 0$:
\[
\lambda_{c,h,p}
\;=\; \lambda_{c,\,h-h,\,R_{-\theta_h}p}
\;=\; \lambda_{c,\,0,\,R_{-\theta_h}p}.
\]
Defining $f_c(q) := \lambda_{c,0,q}$ gives
$\lambda_{c,h,p} = f_c(R_{-\theta_h}p)$, which is
Equation~\ref{eq:orbit_solution}.

What the proof shows is that the orbit-invariance constraint removes exactly one
of the two indices: the field is determined freely on the slice $h = 0$ and is
then forced everywhere else by rotating that slice. This is why the admissible
family is as large as the set of spatial profiles, and why collapsing it to
``independent of $h$'' both loses admissible fields and admits inadmissible
ones.

\section{Proof for Theorem \ref{thm:main}}\label{proof:thm:main}
Fix $g_a \in G$. We compare the network's hidden state when run on the
\emph{rotated} input $\tilde x_t := \rot(g_a) x_t$ against the rotated
hidden state of the network run on the \emph{original} input $x_t$.
Quantities computed from $\tilde x_t$ are denoted with primes
($\tilde i_t, v'_t, s'_t$); those computed from $x_t$ are unprimed
($i_t, v_t, s_t$). By definition $\tilde i_t = \GConv(\tilde x_t) = \GConv(\rot(g_a) x_t)$.
Lemma~\ref{lem:conv} applied at every time step gives:
\begin{equation}
\label{eq:proof_step1}
\tilde i_t = \rot(g_a)\, i_t
\quad \text{for every } t = 1, 2, \dots, T.
\end{equation}
We claim that for every $t \in \{0, 1, \dots, T\}$:
\begin{equation}
\label{eq:induction}
v'_t = \rot(g_a)\, v_t
\quad \text{and} \quad
s'_t = \rot(g_a)\, s_t.
\end{equation}
We proceed to prove this claim in 2 cases:
\begin{itemize}
    \item \textbf{Base case ($t = 0$):} By the initial-condition convention in Equation~\ref{eq:lif_state} we have $v_0 = s_0 = 0$ for the original network, and likewise $v'_0 = s'_0 = 0$ for the rotated-input network. Since $\rot(g_a)\, 0 = 0$ (the zero feature map is fixed by every group action), we obtain $v'_0 = \rot(g_a)\, v_0$ and $s'_0 = \rot(g_a)\, s_0$. Hence Equation~\ref{eq:induction} at $t=0$ holds.
    \item \textbf{Other cases ($t > 0$):} We compute $v'_t$ from the LIF
    recurrence:
    \begin{equation}
    \label{eq:proof_lifprime}
    \begin{aligned}
        v'_t[c,h,p] = & \, \underbrace{\beta_{c,h,p}\, v'_{t-1}[c,h,p](1 - s'_{t-1}[c,h,p])}_{\text{leak }\&\text{ reset}} + \underbrace{\tilde i_t[c,h,p]}_{\text{input current}}.
    \end{aligned}
    \end{equation}
    We process this expression in five sub-steps:
    \begin{enumerate}
        \item Substitute Equation~\ref{eq:induction} and \ref{eq:proof_step1} into Equation~\ref{eq:proof_lifprime}:
        \begin{equation}
        \label{eq:proof_subst}
        \begin{aligned}
            v'_t[c,h,p] = &\, \beta_{c,h,p} \left(\rot(g_a)\, v_{t-1}[c,h,p]\right) \cdot\left(1 - \rot(g_a)\, s_{t-1}[c,h,p]\right) + \rot(g_a)\, i_t[c,h,p].\\
        \end{aligned}
        \end{equation}
        \item The constant feature map $\Ind$ (i.e.\ value $1$ everywhere on
        $\{0,\ldots,n-1\}\times\mathcal{P}$) satisfies $\rot(g_a)\,\Ind = \Ind$,
        because every entry equals $1$ and a permutation of indices maps $1$ to
        $1$. Combined with (R2) of Lemma~\ref{lem:rho}, this gives
        \begin{equation}
        \label{eq:proof_oneminus}
        \begin{aligned}
            & 1 - \rot(g_a)\, s_{t-1}[c,h,p]\\
        & =
        \rot(g_a)\,\Ind - \rot(g_a)\, s_{t-1}[c,h,p]\\
        & =
        \rot(g_a)\,(1 - s_{t-1}[c,h,p]).\\
        \end{aligned}
        \end{equation}
        \item Substituting Equation~\ref{eq:proof_oneminus}
        into Equation~\ref{eq:proof_subst} and applying (R3) of Lemma~\ref{lem:rho} to
        the elementwise product:
        \begin{equation}
            \begin{aligned}
            v'_t[c,h,p] & =  \beta_{c,h,p}\left(\rot(g_a)\, v_{t-1}[c,h,p]\right) \cdot \left(\rot(g_a)\,(1 - s_{t-1}[c,h,p])\right)+ \rot(g_a)\, i_t[c,h,p]\\
            & =  \beta_{c,h,p} \left[\rot(g_a) \left(\,v_{t-1}[c,h,p] \odot (1 - s_{t-1}[c,h,p])\,\right)\right] + \rot(g_a)\, i_t[c,h,p].\\
            \end{aligned}
            \label{eq:proof_R3}
        \end{equation}
        \item The leak field $\beta$ is orbit invariant by hypothesis, so by (R5)
        of Lemma~\ref{lem:rho} it passes through $\rot(g_a)$:
        \begin{equation}
        \label{eq:proof_R1}
        \begin{aligned}
            & \beta_{c,h,p} \left[\rot(g_a)\left(\,v_{t-1}[c,h,p] \odot (1 - s_{t-1}[c,h,p])\,\right)\right]\\
            & =\rot(g_a)\left(\,\beta_{c,h,p} v_{t-1}[c,h,p] \odot (1 - s_{t-1}[c,h,p])\,\right).
        \end{aligned}
        \end{equation}
        Written out at the index, the middle step of (R5) is the substitution
        $\beta_{c,h,p} = \beta_{c,\,h-a,\,R_{-\theta_a}p}$, which is
        Equation~\ref{eq:orbit}. This is the first of the two places where the
        hypothesis is consumed. A field that violates Equation~\ref{eq:orbit}
        leaves behind the residue computed in
        Equation~\ref{eq:naive_residue}.
        \item Applying (R2) of Lemma~\ref{lem:rho} to the
        sum in Equation~\ref{eq:proof_R3}:
        \begin{equation}
        \label{eq:proof_R2}
        \begin{aligned}
            v'_t[c,h,p] & = \rot(g_a)\left(
                   \beta_{c,h,p}\, v_{t-1}[c,h,p] \odot (1 - s_{t-1}[c,h,p])\right) + \rot(g_a) \, i_t[c,h,p]\\
                & = \rot(g_a)\left(\beta_{c,h,p} v_{t-1}[c,h,p]\,(1 - s_{t-1}[c,h,p])  + i_t[c,h,p]\right).
        \end{aligned}
        \end{equation}
        The expression inside the outermost $\rot(g_a)$ on the right-hand side is
        exactly the LIF recurrence in Equation~\ref{eq:lif_state} for $v_t$. Therefore:
        \begin{equation}
        \label{eq:proof_v}
        v'_t = \rot(g_a)\, v_t.
        \end{equation}
        For the spike output we compare the two indicators index by index. This
        is the second and last place where the hypothesis is consumed:
        \begin{equation}
        \begin{aligned}
            s'_t[c,h,p]
            & = \Ind\left[v'_t[c,h,p] \geq v_{\mathrm{th},c,h,p}\right]
            && \text{(definition of $s'$)}\\
            & = \Ind\left[\left(\rot(g_a) v_t\right)[c,h,p] \geq v_{\mathrm{th},c,h,p}\right]
            && \text{(by Equation~\ref{eq:proof_v})}\\
            & = \Ind\left[v_t\left[c,h-a,R_{-\theta_a}p\right] \geq v_{\mathrm{th},c,h,p}\right]
            && \text{(definition of the action)}\\
            & = \Ind\left[v_t\left[c,h-a,R_{-\theta_a}p\right] \geq v_{\mathrm{th},\,c,\,h-a,\,R_{-\theta_a}p}\right]
            && \text{(Equation~\ref{eq:orbit} for $v_{\mathrm{th}}$)}\\
            & = s_t\left[c,h-a,R_{-\theta_a}p\right]
            && \text{(definition of $s$)}\\
            & = \left(\rot(g_a)\, s_t\right)[c,h,p]
            && \text{(definition of the action)}.
        \end{aligned}
        \label{eq:proof_s}
        \end{equation}
        When $v_{\mathrm{th}}$ is a global constant the same chain collapses to a
        single application of (R4) of Lemma~\ref{lem:rho}.
        Equations~\ref{eq:proof_v} and~\ref{eq:proof_s} together establish
        Equation~\ref{eq:induction}, closing the induction.
     \end{enumerate}
\end{itemize}

The output of $L$ at time $t$ is $s_t$ on the original input and $s'_t$ on
the rotated input. By Equation~\ref{eq:induction}, $s'_t = \rot(g_a)\, s_t$ for every $t \in \{1, \dots, T\}$. Stacking along the time axis (on which
$\rot(g_a)$ acts trivially) gives
$L(\rot(g_a)\, x_{1:T}) = \rot(g_a)\, L(x_{1:T})$, which
is Equation~\ref{eq:thm_claim}.

Orbit invariance was used in exactly two places, Equation~\ref{eq:proof_R1} for the leak and Equation~\ref{eq:proof_s} for the threshold, and nowhere else. Three readings
follow. First, sharing across \emph{channels} is never needed, because the
group action does not touch the channel index; per-channel $\beta_c$ and
$v_{\mathrm{th},c}$ are admissible. Second, the two parameters enter through
structurally different steps, a multiplicative one and a comparison, which is
why Section~\ref{sec:numval} finds that they fail on different time scales: the
threshold corrupts the very first indicator whereas the leak is inert at $t=1$
because $v_0 = 0$ and only accumulates from $t=2$. Third, nothing in the
argument refers to the algebraic form of the LIF update beyond the fact that it
is applied elementwise, which is the observation
Proposition~\ref{prop:general_neuron} formalises.

\section{Proof for Proposition \ref{prop:general_neuron}}\label{proof:prop:general}

\noindent\textbf{Sufficiency.} We first record the multi-argument version of
(R4) of Lemma~\ref{lem:rho}. Let $\Xi : \R^{d} \times \R \to \R^{d'}$ be any
fixed map and let $U, I$ be feature maps of the appropriate value type. Applying
$\Xi$ elementwise and then relabelling indices gives the same result as
relabelling first, because the two operations act on disjoint axes:
\[
\left(\rot(g_a)\,\Xi(U, I)\right)[c,h,p]
= \Xi\!\left(U[c,h-a,R_{-\theta_a}p],\, I[c,h-a,R_{-\theta_a}p]\right)
= \Xi\!\left(\rot(g_a)U,\, \rot(g_a)I\right)[c,h,p].
\]
Now fix $g_a \in G$ and write primed quantities for the run on the rotated
input. By Lemma~\ref{lem:conv}, $\tilde i_t = \rot(g_a) i_t$ for every $t$. We
claim $u'_t = \rot(g_a) u_t$ for every $t \geq 0$ and prove it by induction.

\noindent\emph{Base case.} $u_0$ is constant over $(h,p)$ by hypothesis, and
every constant feature map is fixed by the action, so
$u'_0 = u_0 = \rot(g_a) u_0$.

\noindent\emph{Inductive step.} Assume $u'_{t-1} = \rot(g_a) u_{t-1}$. Then
\[
\begin{aligned}
u'_t[c,h,p]
 &= \Psi\!\left(u'_{t-1}[c,h,p],\, \tilde i_t[c,h,p];\, \lambda_{c,h,p}\right) \quad \text{(recursion on the rotated run)}\\
 &= \Psi\!\left(u_{t-1}\left[c,h{-}a,R_{-\theta_a}p\right],\, i_t\left[c,h{-}a,R_{-\theta_a}p\right];\, \lambda_{c,h,p}\right)\\
  &\qquad \text{(inductive hypothesis and $\tilde i_t = \rot(g_a) i_t$)}\\
 &= \Psi\!\left(u_{t-1}\left[c,h{-}a,R_{-\theta_a}p\right],\, i_t\left[c,h{-}a,R_{-\theta_a}p\right];\, \lambda_{c,\,h-a,\,R_{-\theta_a}p}\right)
  \quad \text{(Equation~\ref{eq:orbit})}\\
 &= u_t\left[c,h{-}a,R_{-\theta_a}p\right]
  \quad \text{(recursion on the original run)}\\
 &= \left(\rot(g_a) u_t\right)[c,h,p]
  \quad \text{(definition of the action)} .
\end{aligned}
\]
The identical chain with $\Theta$ in place of $\Psi$ gives
$s'_t = \rot(g_a) s_t$, and stacking over $t$ yields the claim.

\noindent\textbf{Necessity.} Suppose $\lambda$ violates
Equation~\ref{eq:orbit}, so that
$\lambda_{c,h,p} \neq \lambda_{c,\,h-a,\,R_{-\theta_a}p}$ for some
$(c,h,p,a)$. By the assumption that $\Psi$ or $\Theta$ depends non-trivially on
$\lambda$, there is a state-input pair $(u, i)$ at which the two parameter
values produce different outputs. Choosing an input sequence whose lifted
current realises that pair at the index in question, which is possible whenever
$\GConv$ is surjective onto a neighbourhood of the relevant current value,
reproduces the single-index discrepancy exhibited for the LIF case in
Appendix~\ref{apd:naive_fails}.

What this proof shows is that the equivariance of a spiking layer is a statement
about \emph{where the parameters live}, not about the neuron model. Any state
recursion applied independently at each $(c,h,p)$, of any state dimension $d$
and with any nonlinearity, inherits Theorem~\ref{thm:main} unchanged as long as
its parameter field satisfies Equation~\ref{eq:orbit}. The corollary that
matters in practice is negative: the guarantee is lost the moment a neuron
couples across the group or spatial axes with a non-equivariant operator, and no
choice of parameter sharing can restore it.

\section{Proof for Lemma \ref{lem:gmax}}\label{proof:lem:gmax}
By Equation~\ref{eq:groupaction}, $(\rho(g_a)\bar S)[c, h, p] = \bar S[c, h - a \bmod n, R_{-\theta_a} p]$. Maximising over $h$ on the left equals maximising over the relabelled index $h' := h - a \bmod n$ on the right, which is the same range, so $\max_h (\rho(g_a)\bar S)[c, h, p]
   = \max_{h'} \bar S[c, h', R_{-\theta_a} p]$.
This makes $M$ \emph{equivariant} on the spatial axes (a rotated input
gives a spatially rotated $M$). Spatial GAP then averages over $p$, and
$\sum_p f(R_{-\theta_a} p) = \sum_p f(p)$ for any rotation that preserves
$\mathcal{P}'$, giving full invariance of $z$.

What the proof shows is that invariance needs two separate symmetric
reductions: the maximum over $h$ removes the cyclic shift of the group axis,
and the average over $p$ removes the spatial rotation. Either one alone
leaves an equivariant, not an invariant, feature, and the second requires the
final spatial grid $\mathcal{P}'$ to be closed under the rotation, which
holds exactly for $C_4$ and $D_4$ on a square grid.

\section{Proof for Proposition \ref{prop:naive_fails} and analysis of the naive construction}
\label{apd:naive_fails}

We now make precise the claim that simply chaining a $G$-equivariant
convolution with a per-element LIF neuron \emph{does not} yield an
equivariant layer. The argument runs in the same notation as established in
Lemma~\ref{lem:rho}. Consider replacing the constant $\beta$ and threshold $v_{\mathrm{th}}$ in
the LIF recurrence in Equation~\ref{eq:lif_state} by parameters that are allowed to
vary across \emph{every} feature index, including the group axis:
\begin{equation}
\label{eq:naive_lif}
\begin{aligned}
    & \begin{aligned}
       v_t[c,h,p] =\; & \beta_{c,h,p}\,v_{t-1}[c,h,p]\,\left(1 - s_{t-1}[c,h,p]\right) + i_t[c,h,p]\\
    \end{aligned},\\
    & s_t[c,h,p] = \Ind\left[v_t[c,h,p] \geq v_{\mathrm{th},c,h,p}\right].
\end{aligned}
\end{equation}
This is the form taken by any SNN that learns per-neuron time constants and thresholds, instantiated on a feature tensor that has been lifted to $G \times \R^2$ by a $G$-equivariant
convolution. Let $L^{\mathrm{naive}} := \mathrm{LIF}_{\beta_\bullet, v_{\mathrm{th},\bullet}} \circ \GConv$ denote the resulting layer. We prove
Proposition~\ref{prop:naive_fails}, namely that
\begin{equation}
\label{eq:naive_neq}
L^{\mathrm{naive}}\left(\rot(g_a)\,x_{1:T}\right)
\;=\;
\rot(g_a)\, L^{\mathrm{naive}}(x_{1:T})
\qquad\text{for all } x_{1:T} \text{ and all } g_a \in G
\end{equation}
holds if and only if $\beta$ and $v_{\mathrm{th}}$ are orbit invariant in the
sense of Definition~\ref{def:orbit}. The sufficiency direction is
Theorem~\ref{thm:main}, so only necessity remains, and we obtain it by
localising the first step of the induction that can fail.

\begin{proof}
    We retrace the induction of Theorem~\ref{thm:main}, looking for the first
    sub-step that fails. Suppose for the sake of argument that the induction
    hypothesis of Equation~\ref{eq:induction} (namely $v'_{t-1} = \rot(g_a)\,v_{t-1}$
    and $s'_{t-1} = \rot(g_a)\,s_{t-1}$) is granted. We compute $v'_t$
    exactly as before until the moment the parameters appear.
    
    The substitutions and the application of Lemma~\ref{lem:rho} (R2), (R3) do
    not invoke $\beta$ or $v_{\mathrm{th}}$. So we still arrive at:
    \begin{equation}
    \label{eq:naive_R3}
        v'_t[c,h,p] = \beta_{c,h,p}\,\left(\rot(g_a) v_{t-1}\right)[c,h,p] \cdot \left(1 - \rot(g_a) s_{t-1}\right)[c,h,p] +\left(\rot(g_a) i_t\right)[c,h,p],
    \end{equation}
    which is just Equation~\ref{eq:proof_subst} rewritten at one specific index compared with
    Equation~\ref{eq:proof_R3}. Substituting the action of Equation~\ref{eq:groupaction}:
    \begin{equation}
    \label{eq:naive_unfold}
    \begin{aligned}
        v'_t[c,h,p] \;=\; \beta_{c,h,p}\,
       v_{t-1}\left[c,\, h-a,\, R_{-\theta_a} p\right] \left(1 - s_{t-1}\left[c,\, h-a,\, R_{-\theta_a} p\right]\right) + i_t\left[c,\, h-a,\, R_{-\theta_a} p\right].
    \end{aligned}
    \end{equation}
    In the proof of Theorem~\ref{thm:main}, sub-step 4 moved the orbit-invariant
    field $\beta$ through $\rot(g_a)$ by property (R5). Here that move is
    no longer available. To see the obstruction precisely, we compute what
    $\rot(g_a)\,v_t$ would be at the same index using the \emph{unrotated}
    recurrence~\eqref{eq:naive_lif}:
    \begin{equation}
        \label{eq:naive_target}
        \begin{aligned}
        \left(\rot(g_a)\, v_t\right)[c,h,p]
        & = v_t\left[c,\, h-a,\, R_{-\theta_a} p\right] \\
        & = \beta_{c,\,h-a,\,R_{-\theta_a} p} \cdot v_{t-1}\left[c,\, h-a,\, R_{-\theta_a} p\right] \cdot \left(1 - s_{t-1}\left[c,\, h-a,\, R_{-\theta_a} p\right]\right) + i_t\left[c,\, h-a,\, R_{-\theta_a} p\right].
        \end{aligned}
    \end{equation}
    Comparing Equation~\ref{eq:naive_unfold} and Equation~\ref{eq:naive_target}
    side by side, the input terms agree (both equal
    $i_t[c,h-a,R_{-\theta_a} p]$), but the multiplicative coefficients in
    front of the leak-and-reset term differ:
    \begin{equation}
    \label{eq:naive_residue}
    \begin{aligned}
        \begin{aligned}
            & v'_t[c,h,p] - \left(\rot(g_a)\,v_t\right)[c,h,p]
        \end{aligned} =\; & \left(\beta_{c,h,p} - \beta_{c,\,h-a,\,R_{-\theta_a} p}\right)\cdot \,v_{t-1}\left[c,h-a,R_{-\theta_a}p\right] \cdot \left(1 - s_{t-1}\left[c,h-a, R_{-\theta_a}p\right]\right).
    \end{aligned}
    \end{equation}
    The residue factorises into a parameter mismatch and a state factor. Since
    $v_{t-1}$ and $s_{t-1}$ are determined by the input, and since for any index
    $(c, h-a, R_{-\theta_a}p)$ one can choose an input sequence whose lifted
    current makes $v_{t-1}\left(1 - s_{t-1}\right)$ non-zero at that index, the
    residue in Equation~\ref{eq:naive_residue} vanishes for \emph{all} inputs if
    and only if the parameter mismatch itself vanishes:
    \begin{equation}
    \label{eq:beta_orbit}
    \beta_{c,h,p} \;=\; \beta_{c,\,h-a,\,R_{-\theta_a} p}
    \qquad \text{for every } c,h,p,a,
    \end{equation}
    which is exactly Equation~\ref{eq:orbit}, the orbit-invariance condition of
    Definition~\ref{def:orbit}. By Lemma~\ref{lem:orbit} this is the same as
    saying that $\beta_{c,h,p} = f_c(R_{-\theta_h} p)$ for some spatial profile
    $f_c$.

    It is worth being explicit that Equation~\ref{eq:beta_orbit} is genuinely a
    joint condition on $(h,p)$, and that neither half of it can be dropped:
    \begin{itemize}
        \item \textbf{Independence of $h$ does not imply
        Equation~\ref{eq:beta_orbit}.} Take $C_4$, a single channel, and
        $\beta_{h,p} := \beta_{p}$ a field that depends only on position, say
        $\beta_p = 0.5$ on the upper half of the grid and $\beta_p = 0.9$ on the
        lower half. This is independent of $h$, yet for $a = 1$ and any $p$ in
        the upper half with $R_{-\pi/2}p$ in the lower half we get
        $\beta_{h,p} = 0.5 \neq 0.9 = \beta_{h-1, R_{-\pi/2}p}$, so the residue
        is non-zero and equivariance fails. Per-neuron learnable time constants
        produce exactly such a field.
        \item \textbf{Equation~\ref{eq:beta_orbit} does not imply independence of
        $h$.} Take $C_4$ and any non-radial profile $f$, for instance
        $f(q) = 0.5$ if $q$ lies in the upper half plane and $0.9$ otherwise,
        and set $\beta_{h,p} := f(R_{-\theta_h} p)$. By the sufficiency half of
        Lemma~\ref{lem:orbit} this satisfies Equation~\ref{eq:beta_orbit}, the
        residue vanishes identically, and the layer is equivariant, yet
        $\beta_{0,p} \neq \beta_{1,p}$ at most positions.
    \end{itemize}
    Consequently the condition ``independent of the group index'' is neither
    sufficient nor necessary. It becomes equivalent to
    Equation~\ref{eq:beta_orbit} only under the additional assumption that
    $\beta$ is spatially homogeneous, $\beta_{c,h,p} \equiv \beta_{c,h}$, in
    which case Equation~\ref{eq:naive_residue} reduces to
    \begin{equation}
    \label{eq:naive_residue_h}
    \begin{aligned}
        & \left(\beta_{c,h} - \beta_{c,h-a}\right)\,v_{t-1}\left[c,h-a,R_{-\theta_a}p\right] \cdot \left(1 - s_{t-1}\left[c,h-a,R_{-\theta_a}p\right]\right),
    \end{aligned}
    \end{equation}
    which vanishes for all $h,a$ if and only if $\beta_{c,h}$ is constant in
    $h$. Because a layer is also required to be translation equivariant on the
    spatial axis, which forces spatial homogeneity, the per-channel constant
    $\beta_c$ is the parameterisation that survives both requirements in
    practice. This is the design rule that \GCLIF{} implements, and the reason
    the two conditions are so easily conflated.
    For the spike output, in the equivariant case (Theorem~\ref{thm:main})
    property (R4) of Lemma~\ref{lem:rho} gave us
    $\Ind[\rot(g_a) v_t \geq v_{\mathrm{th}}]
     = \rot(g_a)\,\Ind[v_t \geq v_{\mathrm{th}}]$,
    which used that $v_{\mathrm{th}}$ is the same scalar at every index. With
    position-dependent $v_{\mathrm{th},c,h,p}$, evaluating at $(c,h,p)$:
    \[
        \begin{aligned}
        & s'_t[c,h,p] = \Ind\left[v'_t[c,h,p] \geq v_{\mathrm{th},c,h,p}\right], \\
        & \begin{aligned}
            \left(\rot(g_a)\,s_t\right)[c,h,p] & = s_t[c,h-a,R_{-\theta_a}p]
           \\
           & = \Ind\left[ 
           \begin{aligned}
               & v_t[c,h-a,R_{-\theta_a}p] \geq v_{\mathrm{th},c,h-a,R_{-\theta_a}p}\\
           \end{aligned}\right].
        \end{aligned}
        \end{aligned}
    \]
    Even if the membrane potentials happened to satisfy
    $v'_t[c,h,p] = v_t[c,h-a,R_{-\theta_a}p]$ (which they do not, by the
    preceding paragraph), the two indicators would still disagree whenever the
    membrane lies in the gap:
    \[
    \left[\begin{aligned}
        & \min(v_{\mathrm{th},c,h,p},v_{\mathrm{th},c,h-a,R_{-\theta_a}p}), \max(v_{\mathrm{th},c,h,p},v_{\mathrm{th},c,h-a,R_{-\theta_a}p})
    \end{aligned}\right].
    \]
    Hence the threshold step is equivariant for all inputs if and only if
    $v_{\mathrm{th},c,h,p} = v_{\mathrm{th},\,c,\,h-a,\,R_{-\theta_a}p}$ for
    every $c,h,p,a$, which is Equation~\ref{eq:orbit} applied to
    $v_{\mathrm{th}}$. Combining this with
    Equation~\ref{eq:beta_orbit} establishes the ``only if'' half of
    Proposition~\ref{prop:naive_fails}; the ``if'' half is
    Theorem~\ref{thm:main}.
\qed\end{proof}

What the proof shows is where the two parameters differ. The leak enters
multiplicatively against the previous state, so its mismatch is invisible
whenever that state is zero and thereafter compounds through the recurrence; the
threshold enters a comparison and so corrupts the output at the first time step
at which any neuron is near threshold. Section~\ref{sec:numval} measures both
effects and finds exactly this separation, which is the sharpest available
confirmation that the analysis has identified the right mechanism rather than
merely an inequality.

The error in Equation~\ref{eq:naive_residue} is generic, but to confirm the failure
without any inductive bookkeeping we exhibit it at $t=1$, where the
hypothesis $v_0 = s_0 = 0$ is satisfied trivially on both sides.

Consider $G = C_4$ and a single channel ($C = 1$, so we drop the index
$c$). Take a per-group-index threshold $v_{\mathrm{th},h} = \tau_h$ with $\tau_0 = 0.5$ and $\tau_1 = \tau_2 = \tau_3 = 1.5$, and any uniform $\beta \in (0,1)$. We evaluate at $t=1$, where the leak term is multiplied by $v_0 = 0$ and therefore drops out of the recurrence for every admissible $\beta$; this isolates the threshold issue without having to leave the admissible range $\beta \in (0,1)$. Choose an input $x_1$ such that the lifted
current $i_1 = \GConv(x_1)$ has value $i_1[h_0, p_0] = 1.0$ at exactly one
location $(h_0, p_0) = (0, p_0)$ and zero elsewhere:
\begin{itemize}
    \item \textbf{Original network at $t=1$:}
    Applying Equation~\ref{eq:naive_lif} at $t=1$ gives $v_1 = \beta\,v_0\,(1-s_0) + i_1 = \beta \cdot 0 \cdot 1 + i_1 = i_1$, so $v_1[0, p_0] = 1.0$ and $v_1$ is zero
    elsewhere. The output is
    $s_1[h, p] = \Ind[v_1[h,p] \geq \tau_h]$. At $(0, p_0)$ we have
    $1.0 \geq \tau_0 = 0.5$, so $s_1[0, p_0] = 1$. Everywhere else,
    $s_1 = 0$.
    \item \textbf{Rotated input:}
    Now apply $\rot(g_1)$, the rotation by $\theta_1 = \pi/2$ on the group
    axis. The lifted current of the rotated input is, by Lemma~\ref{lem:conv},
    $\tilde i_1 = \rot(g_1)\,i_1$, which by Equation~\ref{eq:groupaction} satisfies
    \begin{equation}
    \tilde i_1[h, p]
    \;=\; i_1[h-1 \bmod 4,\, R_{-\pi/2}\,p].
    \end{equation}
    The single non-zero entry of $i_1$ at $(0, p_0)$ thus appears in $\tilde
    i_1$ at the position $(h, p)$ satisfying $h - 1 \equiv 0\bmod 4$ and
    $R_{-\pi/2}\,p = p_0$, i.e.\ at $(h, p) = (1,\, R_{\pi/2}\,p_0)$. Hence
    $\tilde v_1[1,\, R_{\pi/2}\,p_0] = 1.0$ and $\tilde v_1 = 0$ elsewhere.
    The output is $\tilde s_1[h,p] = \Ind[\tilde v_1[h,p] \geq \tau_h]$. At
    $(1, R_{\pi/2}\,p_0)$, the threshold is $\tau_1 = 1.5 > 1.0$, so
    $\tilde s_1 = 0$ \emph{everywhere}.
    \item \textbf{Equivariance would require $\tilde s_1 = \rot(g_1)\,s_1$:}
    Which (relabelling the original output by Equation~\ref{eq:groupaction}) is
    $1$ at $(1, R_{\pi/2}\,p_0)$ and $0$ elsewhere. We computed
    $\tilde s_1 = 0$ everywhere. The two outputs differ at
    $(1, R_{\pi/2}\,p_0)$:
    \begin{equation}
    \begin{aligned}
        & \left(L^{\mathrm{naive}}(\rot(g_1) x_{1})\right)[1, R_{\pi/2}\,p_0] = 0,\quad \left(\rot(g_1)\, L^{\mathrm{naive}}(x_{1})\right)[1, R_{\pi/2}\,p_0] = 1.
    \end{aligned}
    \end{equation}
    This is the explicit counter-example claimed in
    Proposition~\ref{prop:naive_fails}.
\end{itemize}

Two features of Equation~\ref{eq:naive_residue} explain why the
failure of equivariance is more serious than a small numerical
perturbation, and motivate \textsc{GC-LIF} as the principled fix. Even if at some time step the membrane residue in Equation~\ref{eq:naive_residue} is
small, on the next step it enters as part of $v'_t$, gets multiplied by
$\beta$, and is added to the next residue. Unrolling for $T$ steps:
\begin{equation}
\label{eq:compound}
    v'_T - \rot(g_a) v_T = \sum_{t=1}^{T} \left(\prod_{u=t+1}^{T} \beta_{\bullet,h,p} \cdot\left(1 - s'_{u-1}[c,h,p]\right)\right)\Delta_t,
\end{equation}
where $\Delta_t$ is the per-step residue at index $(c,h,p)$:
\[
    \begin{aligned}
        \Delta_t \;=\; & \left(\beta_{c,h,p} - \beta_{c,h-a,R_{-\theta_a}p}\right)
    \,v_{t-1}\left[c,h-a,R_{-\theta_a}p\right] \cdot \left(1 - s_{t-1}\left[c,h-a,R_{-\theta_a}p\right]\right),
    \end{aligned}
\]
plus a correction whenever the spike paths
$s'_{t-1}[c,h,p]$ and $s_{t-1}[c,h-a,R_{-\theta_a}p]$
disagree (which they generically do once the residues have grown enough
to push the membrane across threshold at one location but not the other).
The $\Delta_t$ contributions therefore not only persist but interact
non-linearly with the spike non-linearity.

A standard G-CNN classifier head pools over the group axis at the end:
$\hat y = \mathrm{Pool}_h\left(\mathrm{Rate}_T(s_{\bullet,h,\bullet})\right)$
where $\mathrm{Rate}_T$ averages spikes over time. If the layer were
equivariant, $\rot(g_a) s_{1:T}$ would be a permutation of $s_{1:T}$
along $h$, and any permutation-invariant pool ($\max$, $\mathrm{mean}$,
$\sum$) would produce identical readouts on the rotated and unrotated
inputs. With the naive construction, the residue in Equation~\ref{eq:naive_residue}
generally lands at a different spatial location than the corresponding
unrotated spike, so the orientation-pool and the spatial-pool see
\emph{different} totals, destroying the invariance that motivated
building a G-CNN in the first place.

\section{Dihedral group convolution implementation}\label{apd:dihedral_conv}

\subsection{Linear encoding of $D_n$}
We index every group element by a single integer:
\begin{equation}
\label{eq:dih_encode}
\begin{aligned}
    & g = L(f, a) := f \cdot n + a,\\
    &f \in \{0, 1\}, \quad a \in \{0, \dots, n-1\},
\end{aligned}
\end{equation}
so that slots $[0, \dots, n-1]$ hold the $n$ rotations and slots
$[n, \dots, 2n-1]$ hold the $n$ reflections.

\subsection{Multiplication and inverses}
The non-abelian structure of $D_n$ is captured by four product rules:
\begin{equation}
\label{eq:dih_mul}
\begin{aligned}
&r^a \cdot r^b   = r^{a+b}, &
r^a \cdot s r^b = s r^{b-a}, \\
& s r^a \cdot r^b = s r^{a+b}, &
s r^a \cdot s r^b = r^{b-a},
\end{aligned} \pmod n.
\end{equation}
The asymmetry $r^a \cdot sr^b = sr^{b-a}$ (\emph{not} $sr^{a+b}$) follows
from the relation $sr = r^{-1}s$ and is the only place where $D_n$ truly
differs from $C_n \times \{e, s\}$. Inverses are
\begin{equation}
\label{eq:dih_inv}
(r^a)^{-1} \;=\; r^{-a},
\qquad
(s r^a)^{-1} \;=\; s r^a.
\end{equation}
That every reflection is its own inverse is a direct consequence
of Equation~\ref{eq:dih_mul}.

\subsection{Action table}
A $D_n$-equivariant group convolution must permute its input-group axis
according to $g_{\text{out}}^{-1} \cdot h$ for every output slot
$g_{\text{out}}$ and every input slot $h$. We pre-compute this once at
construction time as a $2n \times 2n$ table of integers
\begin{equation}
\label{eq:dih_action}
\mathcal{A}[g_{\text{out}}, h] := L\bigl(g_{\text{out}}^{-1} \cdot h\bigr),
\end{equation}
register it as a non-trainable buffer, and use it as an index tensor in
every forward pass: the input-group permutation needed for output slot
$g_{\text{out}}$ is the row $\mathcal{A}[g_{\text{out}}, :]$.
Equation~\ref{eq:dih_action} also encodes the inverse operator alone ($\mathcal{A}[g, 0] = L(g^{-1})$) which is convenient for sanity checks.

\subsection{Spatial action of a dihedral element}
For each $g = L(f, a) \in D_n$ acting on a kernel
$K \in \R^{C_{\text{out}} \times C_{\text{in}} \times k \times k}$, the
spatial part is:
\begin{equation}
\label{eq:dih_spatial}
s_g(K)
\;=\;
\begin{cases}
\mathrm{rot}_a(K) & \text{if } f = 0, \\[2pt]
\mathrm{flip}_y\bigl(\mathrm{rot}_a(K)\bigr) & \text{if } f = 1,
\end{cases}
\end{equation}
where $\mathrm{rot}_a$ rotates the kernel by $\theta_a = 2\pi a / n$ and
$\mathrm{flip}_y$ negates the vertical coordinate, that is, reflects across
the horizontal axis, consistent with $s$ in Section~\ref{sec:dihedral}. For $n \in \{1, 2, 4\}$ both $\mathrm{rot}_a$ and $\mathrm{flip}_y$ are pixel-exact permutations; for $n > 4$ the rotation is not, which motivates the steerable variant below.

\subsection{Dihedral lifting convolution}
Given a base kernel $K \in \R^{C_{\text{out}} \times C_{\text{in}} \times k \times k}$
and a planar input $x : \R^2 \to \R^{C_{\text{in}}}$, the dihedral
lifting layer is:
\begin{equation}
\label{eq:dih_lift}
    \begin{aligned}
        & (\mathrm{Lift}\, x)[c, g, p] = \sum_{q \in \mathcal{N}}\bigl\langle s_g(K)[c, :, q], x[:, p + q]\bigr\rangle,\quad g \in \{0, \dots, 2n-1\}.
    \end{aligned}
\end{equation}
A single base kernel $K$ is the only learnable parameter, which is the $2n$
output slots are obtained by applying Equation~\ref{eq:dih_spatial} to it.
Equivariance under $\rot(g)$ for every $g \in D_n$ then follows by the
same argument as for $C_n$ (Section~\ref{sec:lift}): the right-hand side
of Equation~\ref{eq:dih_lift} is the standard lifting convolution with the
$D_n$-orbit of $K$.

\subsection{Dihedral group convolution}
For the hidden layers we keep a learnable kernel
$W \in \R^{C_{\text{out}} \times C_{\text{in}} \times 2n \times k \times k}$
indexed by an explicit input-group axis. The kernel for output slot
$g_{\text{out}}$ is
\begin{equation}
\label{eq:dih_gconv_kernel}
W_{g_{\text{out}}}\bigl[c_{\text{out}}, c_{\text{in}}, h, q\bigr]
:=
s_{g_{\text{out}}}\Bigl(W\bigl[c_{\text{out}}, c_{\text{in}}, \mathcal{A}[g_{\text{out}}, h], q\bigr]
\Bigr),
\end{equation}
i.e.\ we first permute the input-group axis using the action
table in Equation~\ref{eq:dih_action} and then apply the spatial part of
$g_{\text{out}}$ to every spatial slice. The convolution itself is:
\begin{equation}
\label{eq:dih_gconv}
(\GConv\, F)\left[ c_{\text{out}}, g_{\text{out}}, p\right] =
\sum_{c_{\text{in}}}\sum_{h=0}^{2n-1}\sum_{q \in \mathcal{N}}
\left(
\begin{aligned}
   & W_{g_{\text{out}}}\left[c_{\text{out}}, c_{\text{in}}, h, q\right]\cdot F\left[c_{\text{in}}, h, p+q\right]
\end{aligned}\right).
\end{equation}
At implementation level the $2n$ output kernels are stacked along the output-channel axis,
\(
W_{\text{flat}} \in \R^{(2n \cdot C_{\text{out}}) \times (2n \cdot C_{\text{in}}) \times k \times k},
\)
so that a single convolution call produces all $2n$ output slots in
parallel.

\subsection{Steerable basis for $n > 4$}
For $n \in \{1, 2, 4\}$ Equations~\ref{eq:dih_lift}, \ref{eq:dih_gconv_kernel} and \ref{eq:dih_gconv}
are pixel-exact. For larger $n$ we replace the discrete kernel with a
continuous steerable parameterisation, identical in form to the $C_n$
case Equation~\ref{eq:steerable} but with the angular argument modified by the
reflection bit. Writing the kernel in polar pixel coordinates
$(\rho, \varphi)$:
\begin{equation}
\label{eq:dih_steerable_def}
K_g(\rho, \varphi) =\sum_{m,\ell}\left(
\begin{aligned}
    & \alpha_{m,\ell}\,\psi_m(\rho)\,\cos\bigl(\ell \,\varphi_g\bigr) + \beta_{m,\ell}\,\psi_m(\rho)\,\sin\bigl(\ell \,\varphi_g\bigr)\\
\end{aligned}\right),
\end{equation}
where $\psi_m(\rho) = \exp\bigl(-(\rho - \rho_m)^2 / 2\sigma^2\bigr)$
are Gaussian radial profiles and the angular argument is:
\begin{equation}
\label{eq:dih_steerable_angle}
\varphi_g
\;=\;
\begin{cases}
\varphi - \theta_a & \text{if } g = L(0, a) = r^a, \\
-\varphi - \theta_a      & \text{if } g = L(1, a) = sr^a.
\end{cases}
\end{equation}
The single sign flip on $\varphi$ in the reflection case is what enforces
$D_n$-equivariance in closed form: it is exactly the substitution
$(\rho, \varphi) \mapsto (\rho, -\varphi)$ induced by $s$.
Crucially, the trainable coefficients
$(\alpha_{m,\ell}, \beta_{m,\ell})$ in Equation~\ref{eq:dih_steerable_def} are
\emph{shared across all $2n$ slots}; the slot index $g$ enters only
through the evaluation pattern of Equation~\ref{eq:dih_steerable_angle}. The
parameter count is therefore identical to the $C_n$ steerable basis and the
extra reflection equivariance is free.

\section{Effect of the evaluation pipeline}
\label{apd:pipeline}

All results in the main text use the protocol of Section~\ref{sec:protocol}.
Preliminary experiments used a simpler pipeline, and one of their results
contradicted Lemma~\ref{lem:gmax}. This appendix compares the two pipelines on
the same models and datasets, so that each element of the final protocol can be
judged by its effect on the numbers.

\subsection{The two pipelines}
The preliminary pipeline differed from the final one in four ways:
\begin{enumerate}[label=(\roman*)]
    \item there was no validation split: each model was trained on the full
    training set, and the reported number was its highest test accuracy over the
    $100$ epochs;
    \item the clean and the transformed test accuracies came from separately
    trained networks;
    \item the transformed test set drew one random group element per image in
    every epoch;
    \item transformations were applied after normalisation, and at $45^\circ$
    the pixels leaving the frame were set to zero, that is, to the dataset mean
    intensity, in transformed images only, so that rotated images gained a
    uniform frame that clean images did not have.
\end{enumerate}
Results were averaged over four runs. The final pipeline replaces (i) by
checkpoint selection on a held-out validation split, (ii) and (iii) by a single
checkpoint evaluated on the whole orbit, and (iv) by the disc of
Section~\ref{sec:transform}, applied to every image. The models, their kernel
parameterisations and the training hyperparameters are the same in both
pipelines.

\subsection{Exact groups}

\begin{table*}[t]
\centering
\caption{Mean test accuracy (\%) under the preliminary and the final pipeline (Appendix~\ref{apd:pipeline}) for the clean setting and the exact groups, and the change $\Delta$ = final $-$ preliminary. Final values are those of Tables~\ref{tab:raw} to~\ref{tab:dn_bench}; confidence intervals are given there.}
\label{tab:pipeline_exact}
\resizebox{\textwidth}{!}{\begin{tabular}{@{}llrrrrrrrrrrrr@{}}
\toprule
\multirow{2}{*}{\textbf{Setting}} & \multirow{2}{*}{\textbf{Model}} & \multicolumn{3}{c}{\textbf{MNIST}} & \multicolumn{3}{c}{\textbf{FMNIST}} & \multicolumn{3}{c}{\textbf{KMNIST}} & \multicolumn{3}{c}{\textbf{CIFAR10}} \\
\cmidrule(lr){3-5}\cmidrule(lr){6-8}\cmidrule(lr){9-11}\cmidrule(lr){12-14}
 & & Prelim. & Final & $\Delta$ & Prelim. & Final & $\Delta$ & Prelim. & Final & $\Delta$ & Prelim. & Final & $\Delta$ \\
\midrule
\multirow{4}{*}{Clean} & CNN & 99.21 & 99.20 & $-0.01$ & 91.95 & 91.13 & $-0.82$ & 94.67 & 93.88 & $-0.79$ & 80.88 & 79.93 & $-0.95$ \\
 & SCNN & 98.73 & 98.68 & $-0.05$ & 90.84 & 89.84 & $-1.00$ & 89.49 & 89.20 & $-0.29$ & 76.67 & 75.77 & $-0.90$ \\
 & GCNN & 98.16 & 97.86 & $-0.30$ & 91.82 & 91.37 & $-0.45$ & 91.93 & 91.61 & $-0.32$ & 80.84 & 79.81 & $-1.03$ \\
 & \SGE{} & 95.75 & 95.58 & $-0.17$ & 88.64 & 87.78 & $-0.86$ & 89.57 & 88.30 & $-1.27$ & 73.61 & 73.58 & $-0.03$ \\
\midrule
\multirow{4}{*}{$C_4$, \protect\xmark} & CNN & 46.78 & 43.17 & $-3.61$ & 37.58 & 34.53 & $-3.05$ & 36.89 & 35.28 & $-1.61$ & 48.17 & 46.73 & $-1.44$ \\
 & SCNN & 46.36 & 45.66 & $-0.70$ & 39.22 & 36.70 & $-2.52$ & 35.03 & 33.97 & $-1.06$ & 47.26 & 46.68 & $-0.58$ \\
 & GCNN & 98.08 & 97.86 & $-0.22$ & 91.30 & 91.37 & $+0.07$ & 91.64 & 91.61 & $-0.03$ & 80.02 & 79.81 & $-0.21$ \\
 & \SGE{} & 95.76 & 95.58 & $-0.18$ & 88.62 & 87.78 & $-0.84$ & 80.41 & 88.30 & $+7.89$ & 72.78 & 73.59 & $+0.81$ \\
\midrule
\multirow{4}{*}{$C_4$, \protect\cmark} & CNN & 94.60 & 96.32 & $+1.72$ & 88.08 & 87.72 & $-0.36$ & 80.81 & 79.57 & $-1.24$ & 74.14 & 73.02 & $-1.12$ \\
 & SCNN & 92.76 & 91.82 & $-0.94$ & 84.18 & 83.60 & $-0.58$ & 67.65 & 65.58 & $-2.07$ & 68.15 & 67.03 & $-1.12$ \\
 & GCNN & 98.00 & 97.85 & $-0.15$ & 91.76 & 91.46 & $-0.30$ & 91.65 & 91.62 & $-0.03$ & 80.17 & 79.72 & $-0.45$ \\
 & \SGE{} & 95.91 & 95.37 & $-0.54$ & 88.64 & 87.83 & $-0.81$ & 81.32 & 88.50 & $+7.18$ & 75.80 & 73.33 & $-2.47$ \\
\midrule
\multirow{4}{*}{$D_4$, \protect\xmark} & CNN & 40.84 & 38.40 & $-2.44$ & 37.05 & 33.45 & $-3.60$ & 29.73 & 28.51 & $-1.22$ & 47.87 & 46.59 & $-1.28$ \\
 & SCNN & 41.52 & 39.16 & $-2.36$ & 38.90 & 36.62 & $-2.28$ & 29.16 & 27.65 & $-1.51$ & 47.61 & 46.07 & $-1.54$ \\
 & GCNN & 96.97 & 96.63 & $-0.34$ & 92.81 & 92.78 & $-0.03$ & 92.55 & 92.09 & $-0.46$ & 84.52 & 83.74 & $-0.78$ \\
 & \SGE{} & 93.98 & 93.94 & $-0.04$ & 90.83 & 90.23 & $-0.60$ & 81.69 & 81.44 & $-0.25$ & 77.64 & 77.24 & $-0.40$ \\
\midrule
\multirow{4}{*}{$D_4$, \protect\cmark} & CNN & 94.87 & 93.56 & $-1.31$ & 87.89 & 87.62 & $-0.27$ & 75.53 & 75.15 & $-0.38$ & 74.27 & 73.54 & $-0.73$ \\
 & SCNN & 89.03 & 87.50 & $-1.53$ & 83.81 & 83.46 & $-0.35$ & 63.29 & 61.83 & $-1.46$ & 67.66 & 66.99 & $-0.67$ \\
 & GCNN & 97.07 & 96.80 & $-0.27$ & 92.98 & 92.83 & $-0.15$ & 92.82 & 92.11 & $-0.71$ & 84.61 & 83.69 & $-0.92$ \\
 & \SGE{} & 95.25 & 93.98 & $-1.27$ & 90.97 & 90.04 & $-0.93$ & 81.97 & 81.62 & $-0.35$ & 78.50 & 77.31 & $-1.19$ \\
\bottomrule
\end{tabular}}
\end{table*}

Table~\ref{tab:pipeline_exact} compares the two pipelines for the clean
setting, $C_4$ and $D_4$. Clean accuracy falls under the final pipeline for
every model and dataset, by $0.01$ to $1.27$~pp (median $0.62$~pp). This is the
expected effect of (i): the maximum over epochs of a noisy test accuracy is
biased upward. Most other entries change by about $1$~pp.

One entry changes qualitatively. Under the preliminary pipeline, \SGE{}-$p4$ on
KMNIST had a clean accuracy of $89.57\%$ and a $C_4$ accuracy without
augmentation of $80.41\%$. A difference of $9.16$~pp is impossible for a fixed
invariant network, and under the final pipeline both numbers are $88.30\%$. The
inconsistency came from (ii): the two numbers were produced by different
networks, and (i) and (iii) added selection and sampling noise on top.

\subsection{Group order 8}

\begin{table*}[t]
\centering
\caption{Mean test accuracy (\%) under the preliminary and the final pipeline for group order $8$, and the change $\Delta$ = final $-$ preliminary. \protect\xmark: trained on clean images; \protect\cmark: trained with random group transformations.}
\label{tab:pipeline_n8}
\resizebox{\textwidth}{!}{\begin{tabular}{@{}llrrrrrrrrrrrr@{}}
\toprule
\multirow{2}{*}{\textbf{Setting}} & \multirow{2}{*}{\textbf{Model}} & \multicolumn{3}{c}{\textbf{MNIST}} & \multicolumn{3}{c}{\textbf{FMNIST}} & \multicolumn{3}{c}{\textbf{KMNIST}} & \multicolumn{3}{c}{\textbf{CIFAR10}} \\
\cmidrule(lr){3-5}\cmidrule(lr){6-8}\cmidrule(lr){9-11}\cmidrule(lr){12-14}
 & & Prelim. & Final & $\Delta$ & Prelim. & Final & $\Delta$ & Prelim. & Final & $\Delta$ & Prelim. & Final & $\Delta$ \\
\midrule
\multirow{4}{*}{$C_8$, \protect\xmark} & CNN & 40.44 & 41.02 & $+0.58$ & 29.45 & 25.79 & $-3.66$ & 25.51 & 24.05 & $-1.46$ & 39.20 & 38.96 & $-0.24$ \\
 & SCNN & 40.69 & 41.35 & $+0.66$ & 29.55 & 26.46 & $-3.09$ & 25.42 & 23.65 & $-1.77$ & 37.91 & 37.91 & $+0.00$ \\
 & GCNN & 69.50 & 87.88 & $+18.38$ & 61.89 & 59.31 & $-2.58$ & 56.64 & 56.12 & $-0.52$ & 64.51 & 64.92 & $+0.41$ \\
 & \SGE{} & 77.75 & 85.62 & $+7.87$ & 56.43 & 58.42 & $+1.99$ & 51.75 & 53.31 & $+1.56$ & 58.02 & 59.75 & $+1.73$ \\
\midrule
\multirow{4}{*}{$C_8$, \protect\cmark} & CNN & 95.95 & 95.74 & $-0.21$ & 85.21 & 83.26 & $-1.95$ & 73.49 & 72.86 & $-0.63$ & 72.14 & 67.51 & $-4.63$ \\
 & SCNN & 90.91 & 90.58 & $-0.33$ & 79.67 & 78.46 & $-1.21$ & 59.08 & 58.83 & $-0.25$ & 68.15 & 59.98 & $-8.17$ \\
 & GCNN & 98.50 & 98.30 & $-0.20$ & 91.81 & 90.56 & $-1.25$ & 93.07 & 92.12 & $-0.95$ & 80.17 & 78.37 & $-1.80$ \\
 & \SGE{} & 96.25 & 96.27 & $+0.02$ & 88.04 & 87.26 & $-0.78$ & 81.63 & 80.24 & $-1.39$ & 72.80 & 72.37 & $-0.43$ \\
\midrule
\multirow{4}{*}{$D_8$, \protect\xmark} & CNN & 37.19 & 36.27 & $-0.92$ & 29.62 & 25.18 & $-4.44$ & 23.53 & 21.39 & $-2.14$ & 38.84 & 38.80 & $-0.04$ \\
 & SCNN & 37.20 & 38.13 & $+0.93$ & 29.88 & 25.43 & $-4.45$ & 23.14 & 21.32 & $-1.82$ & 37.62 & 37.79 & $+0.17$ \\
 & GCNN & 71.35 & 86.73 & $+15.38$ & 62.76 & 57.88 & $-4.88$ & 56.95 & 57.51 & $+0.56$ & 65.92 & 67.97 & $+2.05$ \\
 & \SGE{} & 74.65 & 85.39 & $+10.74$ & 57.89 & 58.97 & $+1.08$ & 54.58 & 53.64 & $-0.94$ & 59.31 & 63.36 & $+4.05$ \\
\midrule
\multirow{4}{*}{$D_8$, \protect\cmark} & CNN & 92.65 & 92.06 & $-0.59$ & 84.37 & 83.11 & $-1.26$ & 68.42 & 68.77 & $+0.35$ & 71.65 & 67.44 & $-4.21$ \\
 & SCNN & 86.47 & 86.21 & $-0.26$ & 79.49 & 78.44 & $-1.05$ & 55.48 & 56.76 & $+1.28$ & 67.51 & 59.95 & $-7.56$ \\
 & GCNN & 97.36 & 97.34 & $-0.02$ & 93.02 & 91.80 & $-1.22$ & 93.42 & 93.19 & $-0.23$ & 81.46 & 82.33 & $+0.87$ \\
 & \SGE{} & 94.76 & 94.60 & $-0.16$ & 90.49 & 89.38 & $-1.11$ & 82.90 & 82.14 & $-0.76$ & 73.96 & 75.72 & $+1.76$ \\
\bottomrule
\end{tabular}}
\end{table*}

Table~\ref{tab:pipeline_n8} compares $C_8$ and $D_8$. Two changes stand out.
Without augmentation the equivariant models gain substantially on MNIST, GCNN by
$18.38$~pp at $C_8$ and $15.38$~pp at $D_8$ and \SGE{} by $7.87$ and
$10.74$~pp, so the preliminary advantage of \SGE{} over GCNN on MNIST ($77.75$
against $69.50\%$ at $C_8$, $74.65$ against $71.35\%$ at $D_8$) disappears
(final: $85.62$ against $87.88\%$ and $85.39$ against $86.73\%$, both
unresolved). With augmentation, the disc removes the image corners, which lowers
the plain models on CIFAR-10 by up to $8.17$~pp. The unequal gains show that the
preliminary boundary artifact did not affect all models equally.

\begin{table}[t]
\centering
\caption{Full-orbit accuracy (\%) at group order $8$ without training augmentation, without and with the disc of Section~\ref{sec:transform}, for GCNN with three activations and for \SGE{}. Mean and $95\%$ confidence interval over four seeds. These are training runs separate from those of Tables~\ref{tab:main} and~\ref{tab:dn_bench}, so the entries with the disc differ from those tables by less than $0.7$~pp.}
\label{tab:artifact}
\resizebox{\linewidth}{!}{\begin{tabular}{@{}llcccc@{}}
\toprule
\multirow{2}{*}{\textbf{Data}} & \multirow{2}{*}{\textbf{Model}} &
\multicolumn{2}{c}{$C_8$} & \multicolumn{2}{c}{$D_8$} \\
\cmidrule{3-6}
 & & no disc & disc & no disc & disc \\
\midrule
MNIST & GCNN (ReLU) & 57.74 $\pm$ 18.77 & 87.46 $\pm$ 0.86 & 71.92 $\pm$ 43.17 & 87.03 $\pm$ 8.16 \\
MNIST & GCNN ($T$-level quantiser) & 61.04 $\pm$ 5.54 & 84.84 $\pm$ 25.45 & 66.62 $\pm$ 6.03 & 84.32 $\pm$ 3.97 \\
MNIST & GCNN (noisy ReLU) & 64.69 $\pm$ 24.04 & 87.76 $\pm$ 5.04 & 68.21 $\pm$ 2.00 & 87.43 $\pm$ 0.29 \\
MNIST & \SGE{} (spiking) & 66.82 $\pm$ 23.07 & 85.41 $\pm$ 18.30 & 63.32 $\pm$ 22.60 & 84.75 $\pm$ 7.98 \\
\midrule
KMNIST & GCNN (ReLU) & 55.02 $\pm$ 10.62 & 56.18 $\pm$ 10.98 & 54.60 $\pm$ 7.43 & 56.99 $\pm$ 4.28 \\
KMNIST & GCNN ($T$-level quantiser) & 50.40 $\pm$ 14.34 & 53.19 $\pm$ 1.26 & 51.94 $\pm$ 16.01 & 53.86 $\pm$ 7.48 \\
KMNIST & GCNN (noisy ReLU) & 54.40 $\pm$ 10.53 & 56.36 $\pm$ 6.73 & 54.29 $\pm$ 5.50 & 58.00 $\pm$ 8.51 \\
KMNIST & \SGE{} (spiking) & 50.59 $\pm$ 3.05 & 53.38 $\pm$ 1.47 & 51.51 $\pm$ 20.17 & 53.33 $\pm$ 7.90 \\
\bottomrule
\end{tabular}}
\end{table}

Table~\ref{tab:artifact} isolates change (iv). Everything follows the final
protocol except the disc, which is either applied or omitted; without the disc,
out-of-frame pixels receive the preliminary fill. The table also contains the
activation ablation mentioned in Section~\ref{sec:cyclic}, which varies only the
activation of GCNN: ReLU; a $T$-level quantiser
$\mathrm{quant}_T(u) = \lfloor T\,\mathrm{clip}(u, 0, 1)\rfloor / T$ with a
straight-through gradient, which reproduces the value discretisation of a
spike-rate readout without temporal dynamics; and ReLU followed, during
training, by uniform noise of width $1/T$, which matches the quantisation step
and controls for regularisation. \SGE{} has both the discretisation and the
recurrence. If discretisation helped at $n = 8$, the quantiser should behave
like \SGE{} and the noisy ReLU should not.

On MNIST, removing the artifact raises accuracy by $18.59$ to $29.72$~pp at
$C_8$ and by $15.11$ to $21.43$~pp at $D_8$, by amounts that depend on the model.
With the artifact, \SGE{} has the highest mean at $C_8$, which reproduces the
preliminary result; without it, the ReLU and noisy-ReLU models have the highest
means, and all four intervals overlap. On KMNIST the artifact changes accuracy
by only $1.2$ to $3.7$~pp, and after its removal the two discretised models,
the quantiser GCNN and \SGE{}, have the lowest means ($53.19$ and $53.38\%$
against $56.18\%$ for ReLU at $C_8$), again unresolved. Neither observation
supports the hypothesis that spike-rate discretisation reduces the effect of the
$45^\circ$ sampling residual; the preliminary advantage is explained by the
boundary artifact.

\end{document}